\documentclass[preprint,12pt]{elsarticle}

\usepackage{amssymb}
\usepackage{amsmath,epsfig}
\usepackage{array}
\usepackage{gensymb}
\usepackage[latin1]{inputenc}
\usepackage{epsfig}
\usepackage{multirow}
\usepackage{makecell}
\usepackage{amssymb,amsfonts,amsmath}
\usepackage{graphicx}
\usepackage{float}
\usepackage{algorithm}
\usepackage{algorithmic}
\usepackage{array}
\usepackage{MnSymbol}
\usepackage[modulo]{lineno}
\usepackage[table,xcdraw]{xcolor}
\usepackage{color}
\usepackage{booktabs}
\usepackage[final]{pdfpages}
\usepackage{graphicx}
\usepackage{diagbox}
\usepackage{multirow}
\usepackage[edges]{forest}
\usepackage{enumitem}
\usepackage{comment}
\usepackage{caption}
\usepackage{subcaption}
\useforestlibrary{linguistics}
\usepackage[scientific-notation=true]{siunitx}

\forestapplylibrarydefaults{linguistics}
\usepackage{fixltx2e}
\DeclareMathOperator*{\argmax}{arg\,max}
\DeclareMathOperator*{\argmin}{arg\,min}
\newcommand{\w}{{\bf w}}

\begin{document}

\begin{frontmatter}


\title{Analysis, Characterization, Prediction and Attribution of Extreme Atmospheric Events with \\Machine Learning: a Review}

\author[inst1]{S. Salcedo-Sanz}\ead{sancho.salcedo@uah.es}
\author[instUPM]{J. P\'erez-Aracil}\ead{jorge.perez.aracil@upm.es}
\author[inst2]{G. Ascenso}\ead{guido.ascenso@polimi.it}
\author[javi1,javi2]{J. Del Ser}\ead{javier.delser@tecnalia.com}
\author[inst3]{D. Casillas-P\'erez}\ead{david.casillas@urjc.es}
\author[instDKRZ]{C. Kadow}\ead{kadow@dkrz.de}
\author[inst1]{D. Fister}\ead{dusan.fister@uah.es}
\author[IGEO]{D. Barriopedro}\ead{davidb@fis.ucm.es}
\author[inst5,IGEO]{R. Garc\'ia-Herrera}\ead{rgarciah@ucm.es}
\author[inst2]{M. Restelli}\ead{marcello.restelli@polimi.it}
\author[inst2]{M. Giuliani}\ead{mateo.giuliani@polimi.it}
\author[inst2]{A. Castelletti}\ead{andrea.castelletti@polimi.it}

\address[inst1]{Department of Signal Processing and Communication, Universidad de Alcal\'a,\\
					Ctra. Madrid-Barcelona, km 33, 28805 Alcal\'a de Henares, Madrid, Spain}           
\address[instUPM]{Department of Computer Systems Engineering, Universidad Polit\'ecnica de Madrid,\\
					Calle de Alan Turing, 28038 Madrid, Spain}
\address[inst2]{Environmental Intelligence Laboratory, Politecnico di Milano, 20113 Milano, Italy}
\address[javi1]{TECNALIA, Basque Research and Technology Alliance (BRTA), 48160 Derio, Spain}
\address[javi2]{University of the Basque Country (UPV/EHU), 48013 Bilbao, Spain}    
            
\address[inst3]{Department of Signal Processing and Communication, Universidad Rey Juan Carlos (URJC),
					Camino del Molino, 5, 28942 Fuenlabrada, Madrid, Spain}
            
\address[instDKRZ]{Deutsches Klimarechenzentrum GmbH, Bundesstra$\beta$e 45a, D-20146 Hamburg, Germany}
            
\address[IGEO]{Instituto de Geociencias (CSIC-UCM),\\
					Facultad de Medicina, Ciudad Universitaria, 28040 Madrid, Spain}
            
\address[inst5]{Department of Earth Physics and Astrophysics, Universidad Complutense de Madrid,\\
				Facultad de CC. F\'isicas, Ciudad Universitaria, 28040 Madrid, Spain}

\begin{abstract}
Atmospheric Extreme Events (EEs) cause severe damages to human societies and ecosystems. The frequency and intensity of EEs and other associated events are increasing in the current climate change and global warming risk. The accurate prediction, characterization, and attribution of atmospheric EEs is therefore a key research field, in which many groups are currently working by applying different methodologies and computational tools. Machine Learning (ML) methods have arisen in the last years as powerful techniques to tackle many of the problems related to atmospheric EEs. This paper reviews the ML algorithms applied to the analysis, characterization, prediction, and attribution of the most important atmospheric EEs. A summary of the most used ML techniques in this area, and a comprehensive critical review of literature related to ML in EEs, are provided. A number of examples is discussed and perspectives and outlooks on the field are provided.
\end{abstract}




\end{frontmatter}


\clearpage
\tableofcontents
\clearpage
\section{Introduction}\label{sec:Intro}
Atmospheric Extreme Events (EEs, either weather or climate-related) gravely impact societies \cite{horton2016review}, causing hundreds of thousand of deaths every year \cite{de2004natural,IPCC_wg2}, and producing important collateral effects, such as migrations \cite{marchiori2012impact,carrico2019extreme}, infrastructure damages \cite{may2013addressing}, transportation problems \cite{trinks2012extreme,stamos2015impact}, and damages to agriculture \cite{ciais2005europe,van2012impacts,lal2012adapting} or ecosystems \cite{seneviratne2012changes,knapp2008consequences,van2013novel,woodward2016effects}.

As the number and intensity of EEs has been increasing in the last few decades (likely as a consequence of climate change processes \cite{mitchell2006extreme,herring2015explaining,grant2017evolution}), so has the number of scientific studies on them. In this context, some classical problems associated with EEs are their analysis \cite{herring2015explaining}, detection \cite{zscheischler2013detection,easterling2016detection}, and causation/attribution to human activities \cite{stott2016attribution,hannart2018probabilities,runge2019inferring,madakumbura2021anthropogenic}. Also, different authors have focused their research in studying compound EEs (combinations of multiple EEs that contribute to societal or environmental risk) \cite{zscheischler2017dependence,zscheischler2020typology,zscheischler2018future,raymond2020understanding}, the relationship of EEs with different processes such as carbon cycle \cite{reichstein2013climate,frank2015effects,van2011drought} or soil moisture \cite{hirschi2011observational,whan2015impact}, and the effects of EEs on economics \cite{chavez2015extreme,ackerman2017worst} and their impact on human systems \cite{zscheischler2014few}, to name just a few.

Different mathematical and computational methods have been used to analyze and forecast EEs, including Numerical Weather Methods (NWM) \cite{lavers2013were,yucel2015calibration,vitart2018sub}, statistical and probability-based methods \cite{ferro2007probability,naveau2020statistical,sapsis2021statistics}, non-linear physics and chaos theory \cite{ghil2011extreme,farazmand2019extreme,chowdhury2022extreme}, and, in the last decade, an important number of Machine Learning (ML) and related techniques, a field with an exponential presence in climate and atmospheric sciences \cite{monteleoni2013climate,cohen2019s2s}, climate change studies \cite{rolnick2019tackling}, and Earth System science in general \cite{reichstein2019deep,karpatne2018machine,camps2019perspective,salcedo2020machine,bonavita2021machine,irrgang2021towards}. 

In the last years, Deep Learning (DL) algorithms, a particularly promising branch of ML, have also been applied to climate science problems \cite{kurth2018exascale,ardabili2019deep}, where they have shown great potential to deal with different EE-related problems \cite{liu2016application,ren2020simplified,qi2020using,fang2021survey}. In this paper, we discuss the most important ML methods applied to atmospheric EE-related problems, including DL approaches. It is possible to classify atmospheric EEs in terms of their physical characteristics and impact for human society and ecosystems. We have chosen a number of atmospheric EEs in terms of their impacts to human societies and ecosystems, to carry out the review of ML methods applied to describe them. In this case we have chosen extreme precipitation and floods, extreme temperatures and heatwaves, droughts, severe weather and low-visibility events. We provide a comprehensive review of the works applying ML algorithms for these EEs problems, and we discuss some case studies and the perspectives of this research area in the near future.

The rest of the paper is structured as follows: the next section will give a theoretical overview of some of the ML algorithms most commonly used when studying EEs. Section \ref{sec:LitREv} presents a comprehensive analysis of existing literature on ML techniques for atmospheric EEs problems. Section \ref{sec:SumTemPred} provides several case studies, while sections \ref{sec:Perspectives} and \ref{sec:Conclusions} provide perspectives and a general outlook on future research.

\section{Methods: Theoretical overview of Machine Learning algorithms}\label{sec:methods}

We start by describing different ensemble-based methods such as random forests (RF), frequently used in the last years in prediction problems related to EEs. We also describe statistical learning algorithms such as Support Vector Machines (SVMs) for classification and regression (SVR), we review the basic concepts of neural networks, including Extreme Learning Machines (ELM), and also some recently proposed Deep Learning (DL) algorithms. We close the section with a brief description of techniques for feature selection in prediction problems.

\subsection{Ensemble methods}

Ensemble methods overcome the (potential) limitations in the predictive performance of a single learning model by relying on the randomized combination of several of them \cite{zhou2012ensemble}.
This paradigm assumes that combinations of several, simple ML models can greatly outperform the performance of a single such model and rival with the robustness or generalization capacity of complex ML, such as the neural networks, which involve a huge number of parameters.
A key advantage of ensemble methods compared to complex single-model methods is that since the models that constitute them have very few parameters, they are fast and easy to train. In recent years, many ensemble methods have been developed, with \emph{bagging} and \emph{boosting} being two of the most studied techniques \cite{gonzalez2020practical}, especially for classification and regression problems.

\subsubsection{Bagging}\label{sec:bagging}
\begin{figure}[!ht]
    \centering
    \includegraphics[draft=false, angle=0,width=10cm]{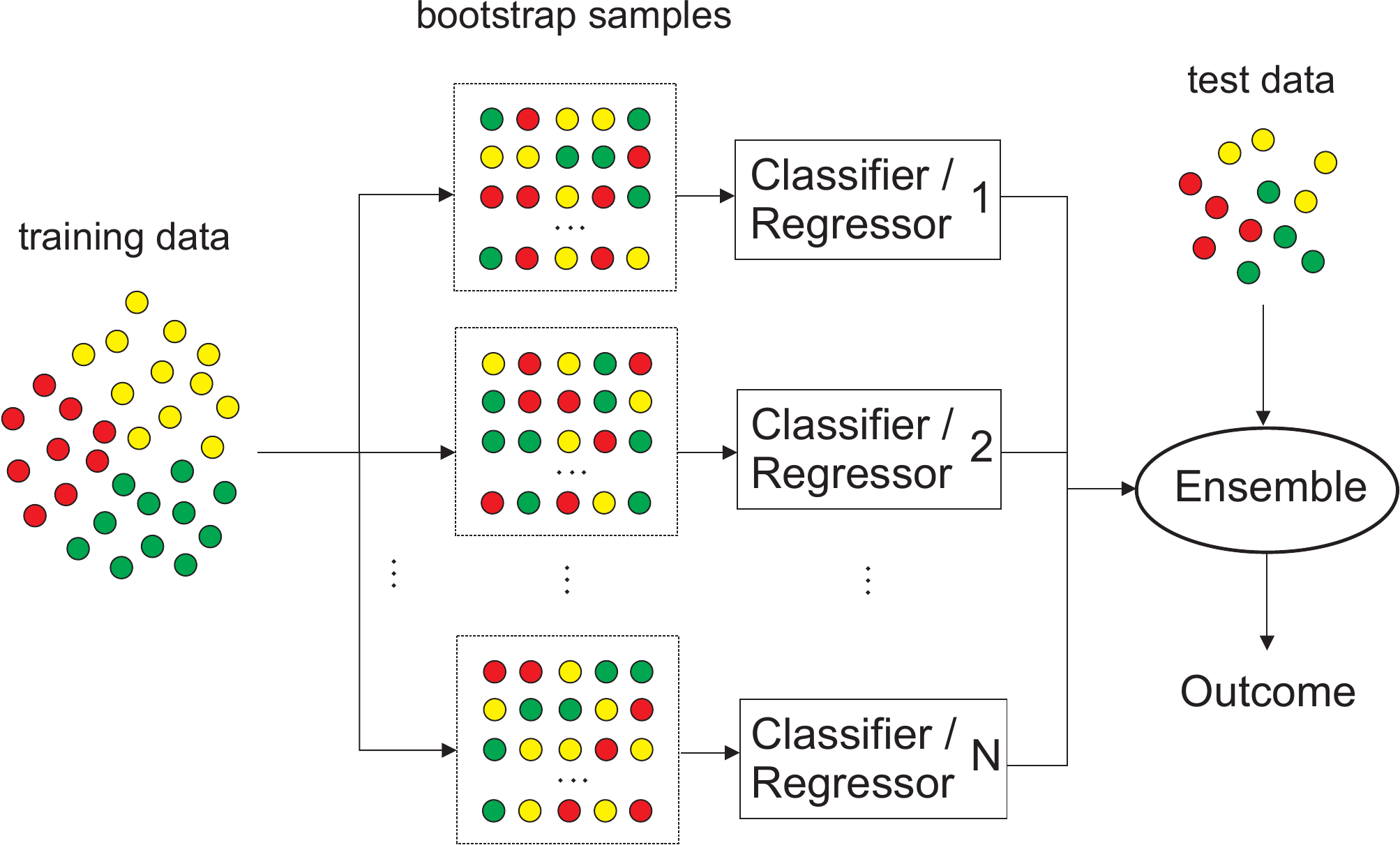}
    \caption{ \label{Bagging_Ejemplo} Diagram of the bagging technique, used for classification or regression problems in ML.}
\end{figure}

We first start with the description of the ensemble technique known as \emph{bootstrap aggregating}, or just {\em bagging}. The basic idea behind bagging is to train a set of simple models and combine their individual predictions as shown in Figure \ref{Bagging_Ejemplo}. 
Bagging also reduces the variance of the ML performance techniques and helps avoid overfitting, which is usually more severe in complex ML methods.
Bagging establishes that all the base ML models which compose the ensemble have the same architecture, which results in same topology, number of input-output variables and number of parameters to train. As an example, a set of decision trees trained with the bagging technique assumes that all trees have the same branches, with the same number of parameters to train and the same input-output variables (see Figure \ref{Bagging_Ejemplo}). The individual models of the ensemble differ in the values that are learned for the model parameters, which are trained with different training sets.

The mathematical description of the bagging technique is as follows: Let $\mathcal{D}=\lbrace(\mathbf{x}_i,y_i)\rbrace_{i=1}^n$ be a given training set of $n$ input-output pairs.
The procedure of bagging, shown in Figure \ref{Bagging_Ejemplo}, generates $N$ {new training sets, of size $n'$, composed of samples from the set $\mathcal{D}$, which can be repeated in each $\mathcal{D}_{i}$. This sampling used for the creation of the sets is known as a {\em bootstrap sample}}. Then, the parameters of $N$ equal models $\lbrace\mathcal{M}_i\rbrace_{i=1}^N$ are learned by training each model $\mathcal{M}_i$ on the respective subset $\mathcal{D}_{i}$.
Finally, the ensemble model combines the individual outputs of each model by averaging their outputs (in the case of regression problems) or by majority voting (if dealing with classification problems) \cite{mohandes2018classifiers}.

Bagging models can be deemed as the simplest way to create ensembles. Note that the base ML models are trained independently with no influence between each other. This property allows to train each model $\mathcal{M}_{i}$ in parallel, which drastically reduces the training time of the ensemble.

Random Forests (RF) \cite{breiman2001random} are among the most commonly applied bagging techniques for classification and regression problems. They specifically use decision or regression trees as learners, and differ from pure bagging techniques in that the topology of the trees is not universally fixed. Trees of the ensemble (the forest) may have different length, topology, or use different input variables, which greatly increases the variability of the learners, but differs the bagging paradigm from a theoretical viewpoint. The main advantage of RFs over traditional bagging is that by adopting slightly different models in their ensemble, the limitations of each are averaged out, resulting in improved generalization capacity \cite{breiman2001random}.

The RF training procedure consists of the following steps.
{Let $\lbrace(\textbf{x}_i,{y}_i)\rbrace_{i=1}^n$ be the training dataset. The main hyperparameters to be adjusted are:} $N$, which is number of estimators (namely, the amount of tree learners composing the forest); and $maxDepth$, which is the maximum number of features to be explored as a node splitting criterion, which is often set to the square root of the number of features. Once these parameters are set, the method works as follows:
\begin{enumerate}[leftmargin=*]
\item Initialize each one of the $N$ decision or regression trees for the classification or regression problem respectively.
\item For each tree $\textbf{T}_t$, select $n_t$ samples with replacement, by using the bootstrapping technique.
\item Only a subset of maximum $maxDepth$ features shall be considered for the construction of each trees.
\item Each tree $\textbf{T}_t$ will give a solution.
\item The ensemble output of the random forest method will be computed by majority voting in the case of classification:
\begin{equation}
    \hat{Y}(\mathbf{x}) = \argmax_l\sum_{t=1}^N[\textbf{T}_t(\mathbf{x})=l].
\end{equation}
or averaging for regression problems: 
\begin{equation}
    \hat{Y}(\mathbf{x}) = \frac{1}{N}\sum_{t=1}^N\alpha_t \textbf{T}_t(\mathbf{x}).
\end{equation}
\end{enumerate}

\subsubsection{Boosting}
\begin{figure}[!ht]
    \centering
    \includegraphics[draft=false, angle=0,width=10cm]{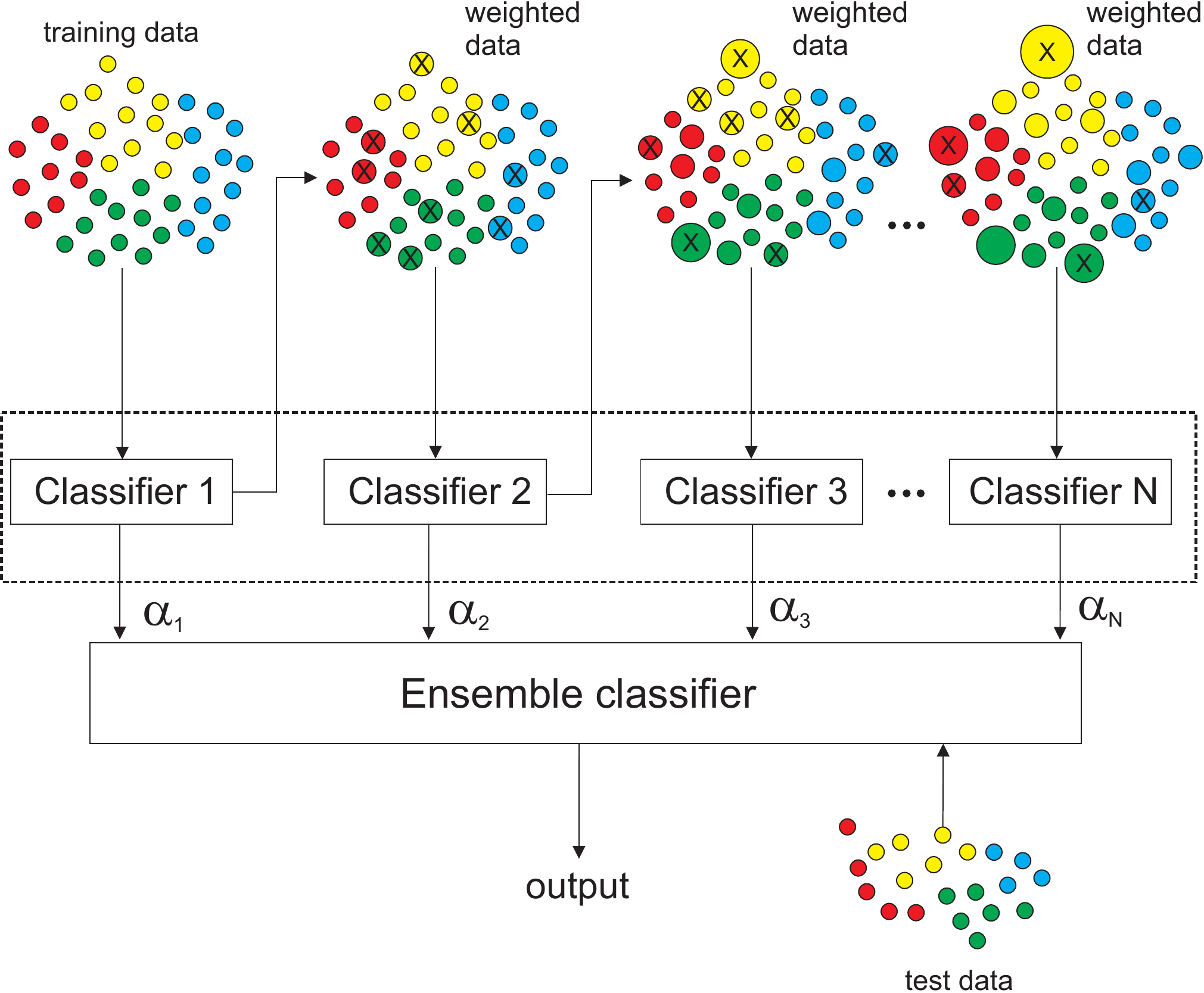}
    \caption{Diagram of the AdaBoost algorithm exemplified for multi-class classification problems. Different size circles stand for samples with more associated weight (w) due to misclassification in the previous step (marked with X).}
    \label{Boosting_Ejemplo} 
\end{figure}

Boosting approaches are an alternative family of ensemble algorithms which perform well in both classification and regression problems \cite{ferreira2012boosting}. Similarly to bagging, boosting follows the learning paradigm of using simple (or ``weak'') ML models (classifiers/regressors), named learners, to form a powerful final model that combines their outputs. Also similarly to bagging, boosting establishes the same topology for all the learners involved in the ensemble (same architecture, number of input-output variables, and number of parameters to train). The most evident difference from bagging lies in the procedure for training the weak learners. In bagging, the weak learners are trained in parallel using different subsets of data $\mathcal{D}_i$ randomly sampled from the whole training dataset $\mathcal{D}$. In boosting, the learners are trained sequentially (see Figure~\ref{Boosting_Ejemplo}). In this way, subsequent learners are dependent on previously trained ones, contrary to the learners in bagging methods. Furthermore, in boosting all the learners use the whole set of training data for computing their parameters, i.e, there is no bootstrap sample step.

Another important difference is that in bagging all input-output pairs are equally weighted to train each learner; each learner equally contributes to determine the final output of the ensemble model.
In boosting, training input-output pairs are weighting according to the accuracy for being predicted by the previous learner (except for the first learner in the queue, which uses the equally weighted samples). Consequently, learners are more specialized as soon as they are placed into the final locations along the queue. Furthermore, the contribution of each learner to the output of the ensemble is usually weighted according to its accuracy, which does not happen in bagging.
This is the general scheme for all boosting methods, but there do exist different boosting strategies depending on the kind of weighting policy applied to each training sample, and/or the output of each learner. 

A widely used boosting technique is Adaptive Boosting (AdaBoost). AdaBoost proposes to train each weak learner in such a way that each learner focuses on the data that was misclassified by its predecessor, so that learners further down the queue iteratively learn to adapt their parameters and achieve better results \cite{ferreira2012boosting,gonzalez2020practical}. Multiple variants of the AdaBoost algorithm exist, starting from the original one \cite{freund1997decision} designed to tackle binary classification problems, regression, or multi-class classification options. Figure \ref{Boosting_Ejemplo} shows an outline of the AdaBoost algorithm for multi-class classification. The pseudocode for AdaBoost can be described as follows:
\begin{enumerate}[leftmargin=*]
\item  {Let $\mathcal{D}=\lbrace(\textbf{x}_i,y_i)\rbrace_{i=1}^n$
be the training dataset.} The first step is to initialise each base learner $\lbrace \textbf{T}_t\;|\;1\leq t \leq N\rbrace$, and assign the set of sample weights $\lbrace {w}_i\;|\;1\leq i \leq n\rbrace$ corresponding to the input-output pairs $\lbrace(\textbf{x}_i,y_i)\rbrace_{i=1}^n$ according to the uniform distribution: ${w}_i = \frac{1}{n}$.
\item  For each base learner $\textbf{T}_t$, the training dataset is used with the distribution of weights ${w}_i$ for training.
\item  After this training process, for each base learner $\textbf{T}_t$, the estimation error $\epsilon_t$ is computed as: \begin{equation}
    \epsilon_t=\sum_{\textbf{T}_t(\textbf{x}_i)\neq {y}_i} \frac{w_i}{\sum_{\textbf{x}_i} w_i},\quad 1\leq i\leq n
\end{equation}
\item  From this error is derived the weight of the current base learner for the ensemble output $\alpha_t$:
\begin{equation}
    \alpha_t = \log \frac{1-\epsilon _t}{\epsilon _t}
\end{equation}
\item  Finally, the distribution of the weights ${w}_i$ corresponding to each $\textbf{x}_i$, which will be used in the next learner, is proportionally adjusted to the probability that a sample is correctly estimated, and inversely proportional to the error of the learner $\epsilon_t$.
\item The final output, provided by the algorithm globally, will be: 
\begin{equation}
    \hat{Y}(\mathbf{x}) = \argmax_l\sum_{t=1}^N[\alpha_t (\textbf{T}_t(\mathbf{x})=l)].
\end{equation}
This final function refers to the boosting method for classification problems, which simply integrates the weighted output of individual learners by voting. In regression problems, the output consists of computing a weighted average of the outputs:
\begin{equation}
    \hat{Y}(\mathbf{x}) = \frac{1}{N}\sum_{t=1}^N\alpha_t \textbf{T}_t(\mathbf{x}).
\end{equation}
\end{enumerate}
The main difference of this algorithm with the multi-class variant AdaBoost.M1 \cite{freund1997decision} is that only the weight values of the correctly classified samples are lowered ($w_i = w_i \frac{\epsilon _t}{1-\epsilon _t}$).

\subsection{Support Vector Machines} \label{svm}

\begin{figure}[h!]
\begin{center}
\includegraphics[width=7cm]{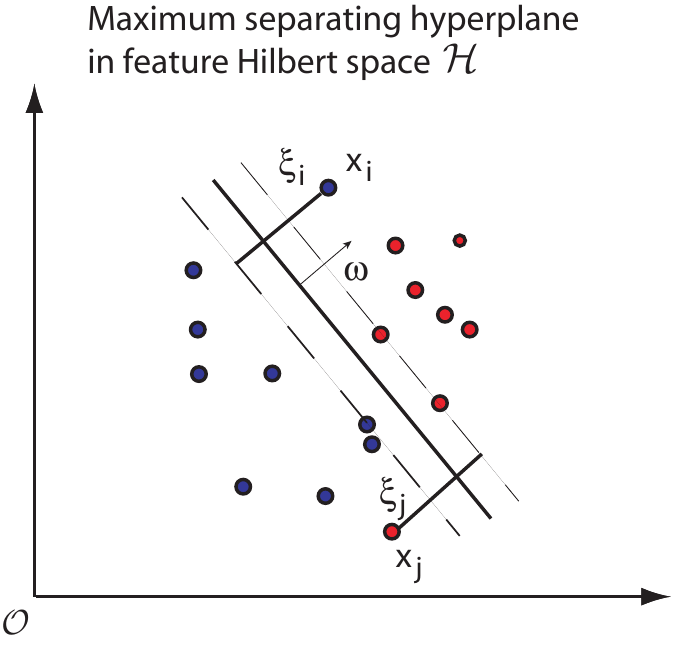}  
\end{center}
\caption{Illustration of the SVM process: Linear decision hyperplanes in a nonlinearly transformed, feature space, where {\em slack} variables $\xi_i$ are included to deal with  errors.}
\label{fig:svc}
\end{figure}

A {\em Support Vector Machine} (SVM) \cite{scholkopf2002learning,scholkopf2000new} is a statistical learning algorithm for classification problems defined as follows: Given a labelled training data set $\{{\bf x}_i,y_i\}_{i=1}^n$, where ${\mathbf x}_{i}\in {\mathbb R}^{N}$ and $y_i\in\{-1,\,+1\}$, and given a nonlinear mapping ${\boldsymbol \phi}(\cdot)$, the SVM method solves the following problem:
\begin{equation}
\min_{\w,\xi_{i},b} \left\{\frac{1}{2}\|\w\|^{2}+C\sum_{i=1}^n \xi_{i}\right\}\label{F2_1}
\end{equation}
constrained to:
\begin{xalignat}{2}
y_{i}\left(\left\langle\boldsymbol{\phi}({\mathbf x}_i),\w\right\rangle +b\right) & \geq 1-\xi_i &  &1\leq i \leq n \label{F2_2}\\
\xi_i & \geq 0 &  &1\leq i\leq n \label{F2_3}
\end{xalignat}
where $\mathbf{w}$ and $b$ define a linear classifier in feature space, and $\xi_i$ are positive slack variables enabling to deal with permitted errors (Figure \ref{fig:svc}). Appropriate choice of nonlinear mapping $\boldsymbol{\phi}$ guarantees that the transformed samples are more likely to be linearly separable in the (higher dimensional) feature space. The regularization hyperparameter $C$ controls the generalization capability of the classifier, and it must be selected by the user. The core problem \eqref{F2_1} is solved using its dual problem counterpart \cite{scholkopf2002learning}, and the decision function for any test vector ${\bf x}_*$ is finally given by
\begin{eqnarray} \label{eq:svc-sol}
f({\bf x}_*) = \operatorname{sgn}\left( \sum_{i=1}^n y_i\alpha_i K({\bf x}_i,{\bf x}_*) + b\right) \label{solution_svm}
\end{eqnarray}
where $\alpha_i$ are Lagrange multipliers corresponding to constraints in \eqref{F2_2}, being the support vectors (SVs) those training samples ${\bf x}_i$ with non-zero Lagrange multipliers $\alpha_i \neq 0$; $K({\mathbf x}_i,{\mathbf x}_*)$ is an element of a kernel matrix ${\bf K}$ \cite{scholkopf2002learning}; and the bias term $b$ is calculated by using the {\em unbounded} Lagrange multipliers as $b = 1/k \sum_{i=1}^k (y_i - \langle\boldsymbol{\phi}({\mathbf x}_i),\w\rangle)$, where $k$ is the number of {\em unbounded} Lagrange multipliers ($0 \leqslant \alpha_i < C$) and $\w = \sum_{i=1}^n y_i \alpha_i \boldsymbol{\phi}({\mathbf x}_i)$ \cite{scholkopf2002learning}.

\subsubsection{Support Vector Regression}\label{sec:SVR}

Support Vector Regression (SVR) \cite{smola2004tutorial} is a well-established algorithm for regression and function approximation problems. SVR takes into account an error approximation to the data, as well as the capability to improve the prediction of the model when a new dataset is evaluated. Although there are several versions of the SVR algorithm, we show the classical model ($\epsilon$-SVR) described in detail in \cite{smola2004tutorial}, which has been used for a large number of problems and applications in science and engineering \cite{salcedo2014support}.

The $\epsilon$-SVR method for regression starts from a given set of training vectors $\{({\bf x}_i,\vartheta_i)\}_{i=1}^N$, where ${\mathbf x}_{i}\in {\mathbb R}^{N}$ and $\vartheta_i\in{ \mathbb R}$, and model the input-output relation as the following general model:
\begin{equation}
\hat{\vartheta}({\bf x})=g({\bf x})+b = {\bf w}^T\phi({\bf x}) + b,
\end{equation}
where ${\bf x}_i$ represents the input vector of predictive variables, $\vartheta_i$ stands for the value of the objective variable $\vartheta$ corresponding to the input vector $\bf{x}_i$ and $\hat{\vartheta}({\bf x})$ represents the model which estimates $\vartheta(\mathbf{x})$. The parameters $(\mathbf{w},b)$ are determined in order to match the training pair set, where the bias parameter $b$ appears separated here. The function $\phi({\bf x})$ projects the input space onto the feature space.
During the training, the algorithms seek those parameters of the model which minimize the following risk function:
\begin{equation}
\label{eq_minfun}
	R[\hat\vartheta] = \frac{1}{2} \left\| {\bf w} \right\|^2 + C \sum_{i=1}^N L\left(\vartheta_i,\hat\vartheta({\bf x}_i)\right),
\end{equation}
where the norm of ${\bf w}$ controls the smoothness of the model and $L\left(\vartheta_i,\hat\vartheta({\bf x}_i)\right)$ stands for the selected loss function. We use the $L^1$-norm modified for the SVR and characterized by the $\epsilon$-insensitive loss function \cite{smola2004tutorial}:
\begin{equation}d
L\left(\vartheta_i,g({\bf x}_i)\right)=\left\{
\begin{array}{l l}
0 & \mbox{if}~~|\vartheta_i-g({\bf x}_i)| \leq \epsilon\\
|\vartheta_i-g({\bf x}_i)|-\epsilon & \mbox{otherwise}.\\
\end{array}
\right.
\end{equation}\\

\begin{figure}[!ht]
\begin{center}
\includegraphics[draft=false, angle=0,width=8cm]{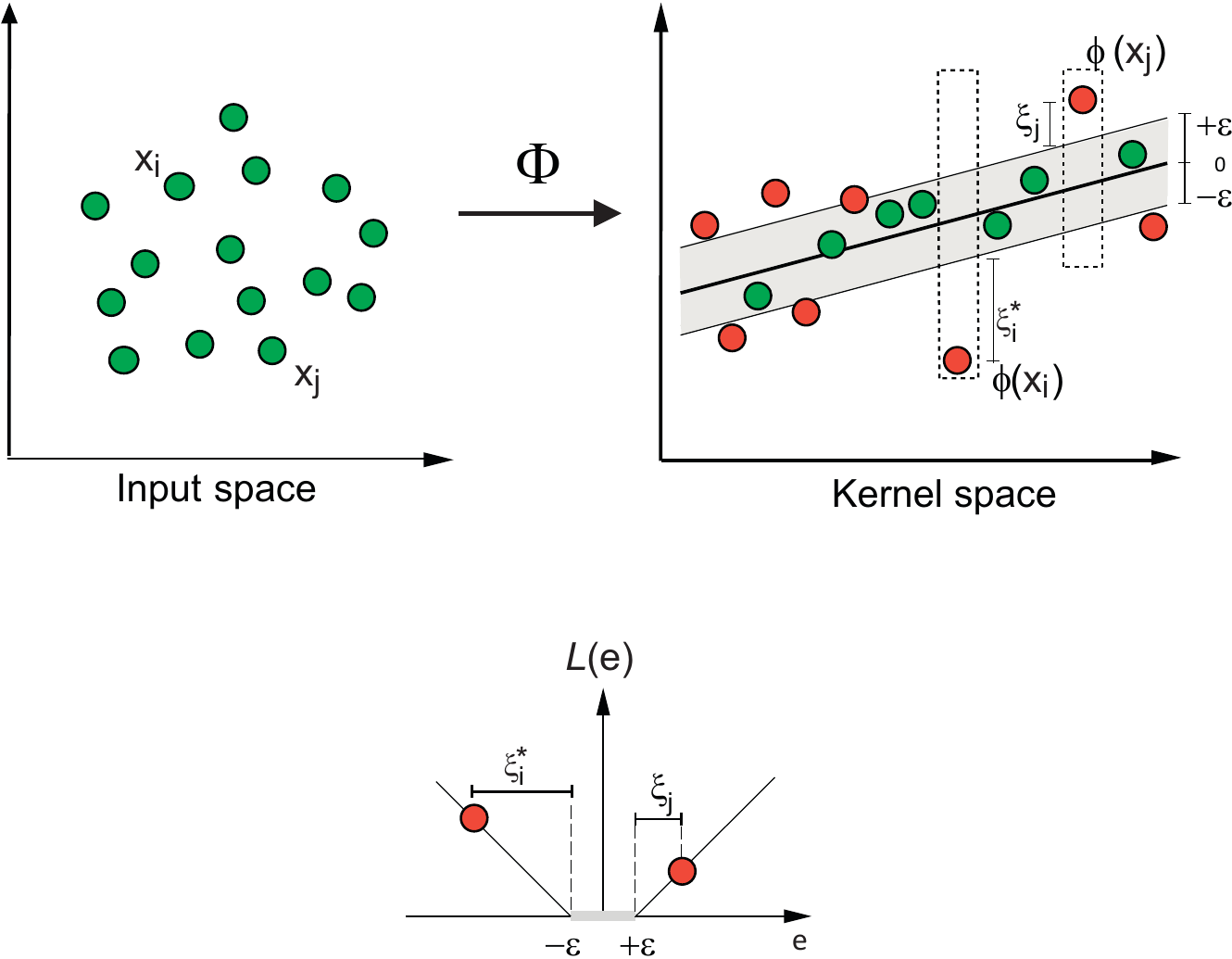}
\end{center}
\caption{ \label{SVR_Ejemplo} Example of a Support-Vector-Regression process for a two-dimensional-regression problem, with an $\epsilon$-insensitive loss function.}
\end{figure}

Figure \ref{SVR_Ejemplo} shows an example of the process of a SVR for a two-dimensional regression problem, with an $\epsilon$-insensitive loss function.

To train this model, it is necessary to solve the following optimization problem \cite{smola2004tutorial}:
\begin{equation}
    \begin{aligned}
        \min_{\mathbf{w},b,\boldsymbol{\xi}} \quad & \frac{1}{2}\lVert \mathbf{w} \rVert+C\sum_{i=1}^{N}{\xi_{i} + \xi_i^*},\\
        \textrm{s.t.} \quad & \vartheta^i - {\bf w}^T\phi({\bf x}_i) - b \leq \epsilon + \xi_i,  &1\leq i \leq N,\\
        &-\vartheta^i + {\bf w}^T\phi({\bf x}_i) + b  \leq \epsilon + \xi_i^*, \quad &1\leq i \leq N,\\
        &\xi_i,\xi_i^* \geq 0, &1\leq i \leq N.
    \end{aligned}
    \label{F2_1*}
\end{equation}

The dual form of this optimization problem is obtained through the minimization of a Lagrange function, which is constructed from the objective function and the problem constraints:
\begin{equation}
    \begin{aligned}
        \max_{{\boldsymbol \alpha},{\boldsymbol \alpha}^*} \quad -\frac{1}{2} \sum_{i,j=1}^N{(\alpha_i-\alpha_i^*)(\alpha_j-\alpha_j^*)K({\bf x}_i,{\bf x}_j)}- \epsilon \sum_{i=1}^N{(\alpha_i+\alpha_i^*)} + \sum_{i=1}^N{\vartheta^i(\alpha_i-\alpha_i^*)},\nonumber
    \end{aligned}
\end{equation}
\begin{equation}
    \begin{aligned}
        \textrm{s.t.} \quad & \sum_{i=1}^l (\alpha_i-\alpha_i^*) = 0, & 1\leq i \leq N,\\
                            & \alpha_i,\alpha_i^* \geq 0,             & 1\leq i \leq N,\\
                            & -\alpha_i,-\alpha_i^* \geq -C,          & 1\leq i \leq N.\\
    \end{aligned}
    \label{F2_2*}
\end{equation}

In the dual formulation of the problem, the function $K({\bf x}_i,{\bf x}_j)$ represents the inner product $\left< \phi({\bf x}_i),\phi({\bf x}_j) \right>$ in the feature space. Any function $K({\bf x}_i,{\bf x}_j)$ may become a kernel function as long as it satisfies the constraints of the inner products. It is very common to use the Gaussian radial basis function:
\begin{equation}
K({\bf x}_i,{\bf x}_j)=\exp(-\gamma \left\|{\bf x}_i-{\bf x}_j\right\|^2).
\end{equation}

The final form of the function $g({\bf x})$ depends on the Lagrange multipliers $\alpha_i,\alpha_i^*$ as:
\begin{equation}
\label{eq:model}
	g({\bf x}) = \sum_{i=1}^N (\alpha_i-\alpha_i^*) K({\bf x}_i,{\bf x}).
\end{equation}

Incorporating the bias, the estimation of the objective function is finally made by the following expression:
\begin{equation}
\hat{\vartheta}({\bf x})=g({\bf x})+b=\sum_{i=1}^N (\alpha_i-\alpha_i^*) K({\bf x}_i,{\bf x})+b \mbox{.}
\end{equation}

\noindent

\subsection{Multi-Layer Perceptrons}\label{sec:MLP}

\begin{figure}[!ht]
\begin{center}
\includegraphics[draft=false, angle=0,width=6cm]{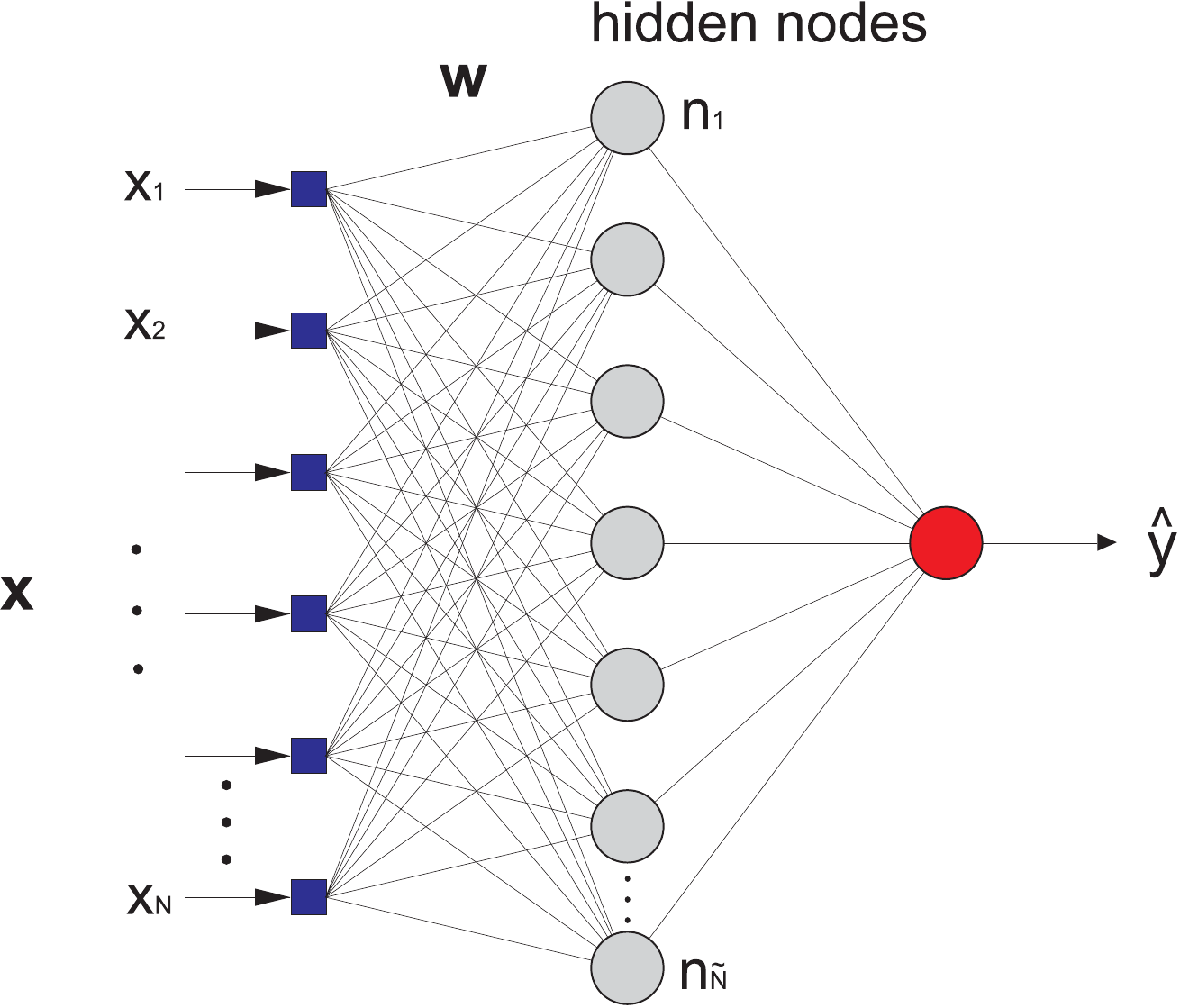}
\end{center}
\caption{ \label{MLP_Ejemplo} Structure of a multi-layer perceptron neural network, with one hidden layer.}
\end{figure}

A multi-layer perceptron (MLP) is a particular class of Artificial Neural Network (ANN), which has been successfully applied to solve a large variety of non-linear problems, mainly classification and regression tasks \cite{haykin2004comprehensive,bishop1995neural}. The multi-layer perceptron consists of an input layer, a number of hidden layers, and an output layer, all of which consist of a number of special processing units called \emph{neurons}. All the neurons in the network are connected to other neurons by means of weighted links (see Figure \ref{MLP_Ejemplo}). In a \emph{feedforward} MLP, the neurons within a given layer are connected to those of the previous layer. The values of these weights are related to the ability of the MLP to learn the problem, and they are learned from a sufficiently long number of examples. The process of assigning values to these weights from labelled examples is known as the training process of the perceptron. The adequate values of the weights minimize the error between the output given by the MLP and the corresponding expected output in the training set. The number of neurons in the hidden layer is also a hyperparameter to be optimized \cite{haykin2004comprehensive,bishop1995neural}.

The input data for the MLP consists of a number of samples arranged as input vectors $\{\mathbf{x}^i\in\mathbb{R}^n\}_{i=1}^N$, with each input vector $\mathbf{x}^i=(x^i_1,\cdots,x^i_n)$. Once an MLP has been properly trained, it can be tested on data it did not see during training to evaluate its performance, in terms of how well the learned weights can transform the given input into a desired output $\vartheta\in\mathbb{R}$. The relationship between the output $\vartheta$ and a generic input signal $\mathbf{x}=(x_1,\cdots,x_n)$ of a neuron is given by:
\begin{equation}
\vartheta({\bf x})=\varphi\left(\sum_{j=1}^n w_j x_j- b\right),
\end{equation}
where $\vartheta$ is the output signal, $x_j$ for $j=1,\ldots,n$ are the input signals, $w_j$ is the weight associated with the $j$-th input, $b$ is the bias term \cite{haykin2004comprehensive,bishop1995neural}, and $\varphi$ is some function chosen based on the type of layer to which it needs to be applied, for example the the logistic function (among other possibilities):
\begin{equation}
\varphi(x)=\frac{1}{1+e^{-x}}.
\end{equation}

The well-known Stochastic Gradient Descent (SGD) algorithm is often applied to train MLPs \cite{rumelhart1986learning}. There are also alternative training algorithms for MLP which have shown excellent performance in different problems, such as the Levenberg-Marquardt algorithm \cite{hagan1994training}, or the ADAM and RMSProp optimizers for training deep versions of the networks \cite{zhang2018improved,zou2019sufficient}.

\subsubsection{Extreme Learning Machines}\label{sec:ELM}
An Extreme Learning Machine (ELM) \cite{Huang06} is a type of training method for multi-layer perceptrons, characterized by being computationally faster than traditional gradient backpropagation \cite{hecht1992theory}. In the ELM algorithm the weights between the inputs and the hidden nodes are set at random, usually by using a uniform probability distribution. Then, the output matrix of the hidden layer  is established and the Moore-Penrose pseudo-inverse of this matrix is computed. The optimal values of the weights belonging to the output layer are directly obtained by multiplying the computed pseudo-inverse matrix with the target (see \cite{Huang12} for details). The ELM obtains competitive results with respect to other classical training methods, while its training computation efficiency overcomes other classifiers or regression approaches such as SVM algorithms or MLPs \cite{Huang12}.

Mathematically, the ELM algorithm considers a training set
$\lbrace(\textbf{x}_i,y_i)\rbrace_{i=1}^n$
to fit the weights $(\beta_k)$ associated with
each hidden node $\tilde{N}$ to optimally compose an output with minimum mean squared error. The training process is according to the following steps:
\begin{enumerate}[leftmargin=*]
\item {The input weights $\textbf{w}_k$ and the bias $b_k$, where $k = 1, \ldots ,\tilde{N}$ are randomly chosen following a uniform distribution with support $[-1,1]$.}
\item {In the second step, the hidden-layer output matrix $H$ is computed as follows:}
\begin{equation}
\textbf{H} = \left[\begin{array}{ccc}
g( \textbf{w}_1 \textbf{x}_1 + b_1) & \cdots & g(\textbf{w}_{\tilde{N}} \textbf{x}_1 + b_{\tilde{N}}) \\
\vdots & \cdots & \vdots \\
g(\textbf{w}_1 \textbf{x}_N + b_1) & \cdots & g(\textbf{w}_{\tilde{N}} \textbf{x}_N + b_{\tilde{N}})
\end{array}
\right]_{\tilde{N}}
\end{equation}
where $g(\cdot)$ is the activation function.
\item The training problem is reduced to a $\boldsymbol\beta$ parameter optimization problem, which can be defined as:
\begin{equation}\label{eq.opt.ELM}
\min\limits_{\boldsymbol\beta} \lVert \textbf{H} \boldsymbol\beta-\textbf{Y}\rVert,
\end{equation}
\item The last step consists in obtaining the output layer weights $\boldsymbol\beta$ by means of the following expression:
\begin{equation}\label{eq.elm}
    \boldsymbol\beta = \textbf{H}^\dagger\textbf{Y}^T,
\end{equation}
{
where $\textbf{Y}^T$ stands for the transpose of the training output vector $\textbf{Y}=[y_1,\ldots,y_n]$ and $\textbf{H}^\dagger$ refers to the Moore-Penrose pseudo-inverse of the hidden-layer matrix $\textbf{H}$ \cite{Huang06}.}
\item Then, the predicted or classified output is obtained as: $\hat{Y}(\mathbf{x}) = \textbf{H} \boldsymbol\beta$.
\end{enumerate}

{The hidden nodes number $\tilde{N}$ can be tuned for improving the ELM performance.}

\subsection{Deep Learning algorithms}

\begin{figure}[h!]
    \centering
    \includegraphics[width=\columnwidth]{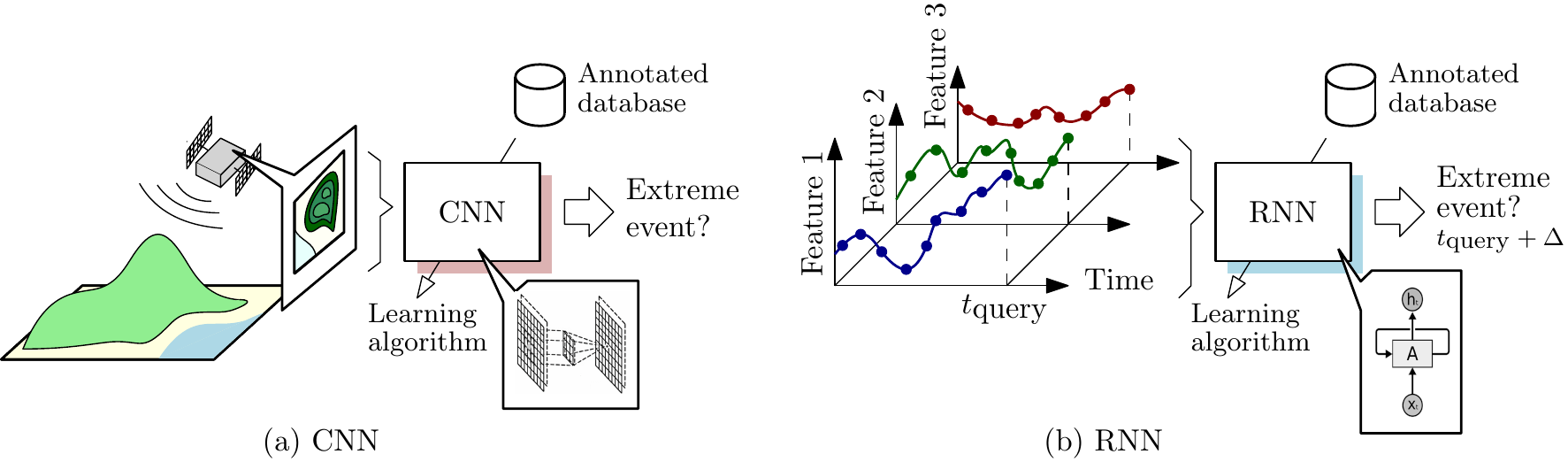}
    \caption{Examples of typical use cases related to extreme atmospheric events that can be tackled by Deep Learning models: (a) CNNs; (b) RNNs.}
    \label{fig:deep_learning_extreme_atmospheric_events}
\end{figure}

When used for predictive modeling, Machine Learning revolves around modeling the statistical correlation between variables with respect to the target variable to be predicted. In problems dealing with spatial and/or temporal data (such as image classification or time series forecasting), such a correlation emerges from the relationship among data points over such domains. As a result, Machine Learning models can be either used in their seminal form to tackle spatio-temporal modeling tasks (by e.g. extracting tabular features from data) or, instead, specialised into archetypes capable of supporting the modeling requirements stemming from such tasks (invariance to spatial transformations of the input or the characterization of long-term correlations over sequential data). Furthermore, continued advances in massive parallel computing and the explosion of non-relational databases containing information of assorted nature (e.g., image, video, audio, text) have spurred research efforts towards the derivation of neural network models of ever-growing modeling complexity, capable of efficiently discovering relevant predictors from highly dimensional data, and endowing mechanisms to meet the requirements mentioned previously. Advances over the past two decades have blossomed into what is now known as \emph{Deep Learning} \cite{lecun2015deep}, which crystallizes in two main neural architectures: Convolutional Neural Networks (CNNs \cite{o2015introduction}) and Recurrent Neural Networks (RNNs \cite{sherstinsky2020fundamentals}). Figure \ref{fig:deep_learning_extreme_atmospheric_events} illustrates two typical applications of these Deep Learning architectures in the context of EEs.

When the correlation is held in the spatial domain, any model should be made invariant with respect to transformations of the input data that should not affect the prediction. This is the case of translational invariance in image classification, by which visual features relevant for the target to be predicted should retain their predictive importance no matter where they are located in the image. The way the human visual cortex operates to satisfy this requisite was the inspiration behind the design of CNNs, which, in their seminal form, comprise a series of hierarchically arranged neural processing layers. Layers closer to the input contain several convolutional neurons (also referred to as convolutional filters or kernels), which extract features from the input data by performing a convolution between the data themselves and the weights at their core. A CNNs for complex modeling tasks may stack several convolutional layers, one after another, so that each layer processes through its filters the output produced by the preceding layer. Some further processing layers can be placed in between convolutional ones, such as pooling layers, which serve to create information bottlenecks that help distil more high-level information while drastically reducing the number of parameters. After the convolutional part of the network, additional layers may be added depending on the application. For instance, in image classification a fully connected multi-layer perceptron is often attached to the end of a CNN to map this output to the target variables to be predicted. Analogously to MLPs, trainable parameters (weights and biases) of the CNN network can be learned by backpropagating error gradients through the network, which also holds for the weights of the convolutional kernels. Since gradients can be computed also for these special neural processing units, their weight values can be adjusted by means of different stochastic gradient descent solvers.

Beyond their benefits in terms of spatial invariance, learnable convolutional layers in CNNs provide several other advantages. First, the fact that gradients can be propagated allows for a massively parallel iterative update of their weights and biases, paving the way for implementations deployable on Graphical Processing Units (GPU) and Tensor Processing Units (TPU). Another advantage of CNNs is the hierarchy of visual features learned by the network, which becomes progressively more specialized for the task at hand as more convolutional layers are stacked on top of each other. This offers a more structured interpretability of the knowledge captured by the layers, which can be disentangled by using deconvolutional filters or local explainability techniques \cite{zhang2018visual}. But perhaps most interestingly, coarse visual features modeled in the first convolutional layers (edges, primitive shapes, etc.) learned on one task can be useful for others. Such tasks could leverage this general-purpose learned knowledge by importing pretrained weights and biases of such layers into their CNN architectures, so that the requirements in terms of learnable parameters or annotated data can be reduced. This simple yet effective knowledge exchange mechanism is referred to as \emph{transfer learning} \cite{zhuang2020comprehensive,weiss2016survey} and has helped the adoption of CNNs in environments with scarcely annotated data or limited computational resources.

Sophisticated CNN architectures nowadays constitute the state-of-the-art for image and video classification modeling tasks, incorporating new ideas that boost even further their performance and/or efficiency. This is the case of capsule networks \cite{hinton2011transforming}, attention mechanisms \cite{vaswani2017attention}, or patch-based learning in visual transformers \cite{han2020survey}. When it comes to efficiency, the inner working of spiking neural networks \cite{gruning2014spiking} has been investigated to alleviate the consumption of computing resources of these models. It is worth noting that the number of trainable parameters in CNNs may amount up to several tens of millions in very deep models, leading to problematically long training times, large storage requirements, and energy consumption footprints \cite{anthony2020carbontracker}. Finally, an important area of research is on the development of interpretability techniques for CNNs, which aim to dissect the knowledge captured by the layers of an already trained CNN \cite{arrieta2020explainable}. The result of this dissection, which can take many forms (e.g., attribution maps, counterfactual explanations, or simplified rule sets) is offered as an interpretable interface for the user to understand how and why the CNN provides its output. We will later elaborate on the plethora of possibilities of explanation techniques for CNNs used in EEs modeling and characterization tasks.\\

Differently from CNNs, RNNs are built for modelling relationships in sequential data, including text and time series. Modeling such correlations requires that the network be capable of modeling, exploiting, and maintaining information (memory) at their neural processing steps, such that long-term relationships over the sequence can be exploited effectively when solving modeling tasks. In RNNs, this is accomplished by formulating a recurrent form of a neural processing unit, in which part of the output of the neuron is fed back to its input to realize a sort of neural memory. This new recurrent formulation of a neuron endows it with the possibility to learn and store information about the past that is relevant for the problem under consideration. For instance, this property of RNNs is key in time series forecasting, where the temporal lags to be predicted can be affected by data occurring far back in time. When RNNs are used for this task, the memory conferred to the neurons permits to model correlations over the sequence at different time scales. As the convolutional filters in a CNN, the parameters controlling how much of the output of a neuron is fed back to its input or stored in the hidden state vector can be learned via gradient backpropagation.
The history of RNNs dates back to the work by Jordan \cite{jordan1997serial} and Elman \cite{elman1990finding}. Thereafter, the well-known Long Short-Term Memory networks (LSTM \cite{hochreiter1997long}) and the more recently proposed Gated Recurrent Units (GRU \cite{cho2014properties}) became the standard in recurrent neural computation. LSTMs rely on several trainable parameters (gates) to control which parts of the sequence flow into the neuron by releasing or retaining information inside the hidden state vectors of neurons. GRU networks can be regarded as a variant of LSTMs that features small architectural modifications that permit to reduce the number of trainable parameters. In both cases, recurrent neural processing units can be arranged in a hierarchical structure comprising several stacked layers, in such a way that correlations are captured at different scales and levels of granularity. 
Several RNN approaches have been proposed in the literature over the years to overcome drawbacks of the training process of these models. Attention mechanisms for instance (also applied in other types of deep networks such as CNN), make networks focus on certain parts of the input when predicting its output, discarding information that is not relevant for that specific input. Similarly, bidirectional RNNs aim at considering future steps of the sequence in the output of the neuron \cite{schuster1997bidirectional}). Recurrent networks that do not hinge on gradient backpropagation have also been developed in recent years, with Reservoir Computing and particularly Echo State Networks \cite{lukovsevivcius2009reservoir, gallicchio2017deep} being at the frontline. Finally, recent studies have emphasized that specialized CNNs for sequence modeling such as Temporal Convolutional Networks (TCN \cite{lea2017temporal}) demonstrate longer and more effectively trained memory capabilities over diverse tasks and datasets, showcasing the potential of convolutional architectures to also address problems over sequential data.

\subsection{Feature selection methods}\label{FS_methods}
For ML-based methods, using irrelevant or redundant features as inputs during training can be detrimental, not only because these additional features would increase the training time, but also because they may hinder their generalisability \cite{Blum97}. In its more general form, the Feature Selection Problem (FSP) for a learning problem from data can be defined as follows: given a set of labelled data samples $\left\lbrace(\mathbf{x}_1,y_1),\ldots,(\mathbf{x}_l,y_l)\right\rbrace$, where ${\mathbf x}_i \in \mathbb{R}^n$ and $y_i \in \mathbb{R}$ (or $y_i \in\{\pm 1\}$ in the case of classification problems), choose a subset of $m$ features ($m<n$), that achieves the lowest error in the prediction of the variable $y_i$.

\begin{figure}[!ht]
\centering
\begin{subfigure}[b]{0.45\linewidth}
    \includegraphics[width=\textwidth]{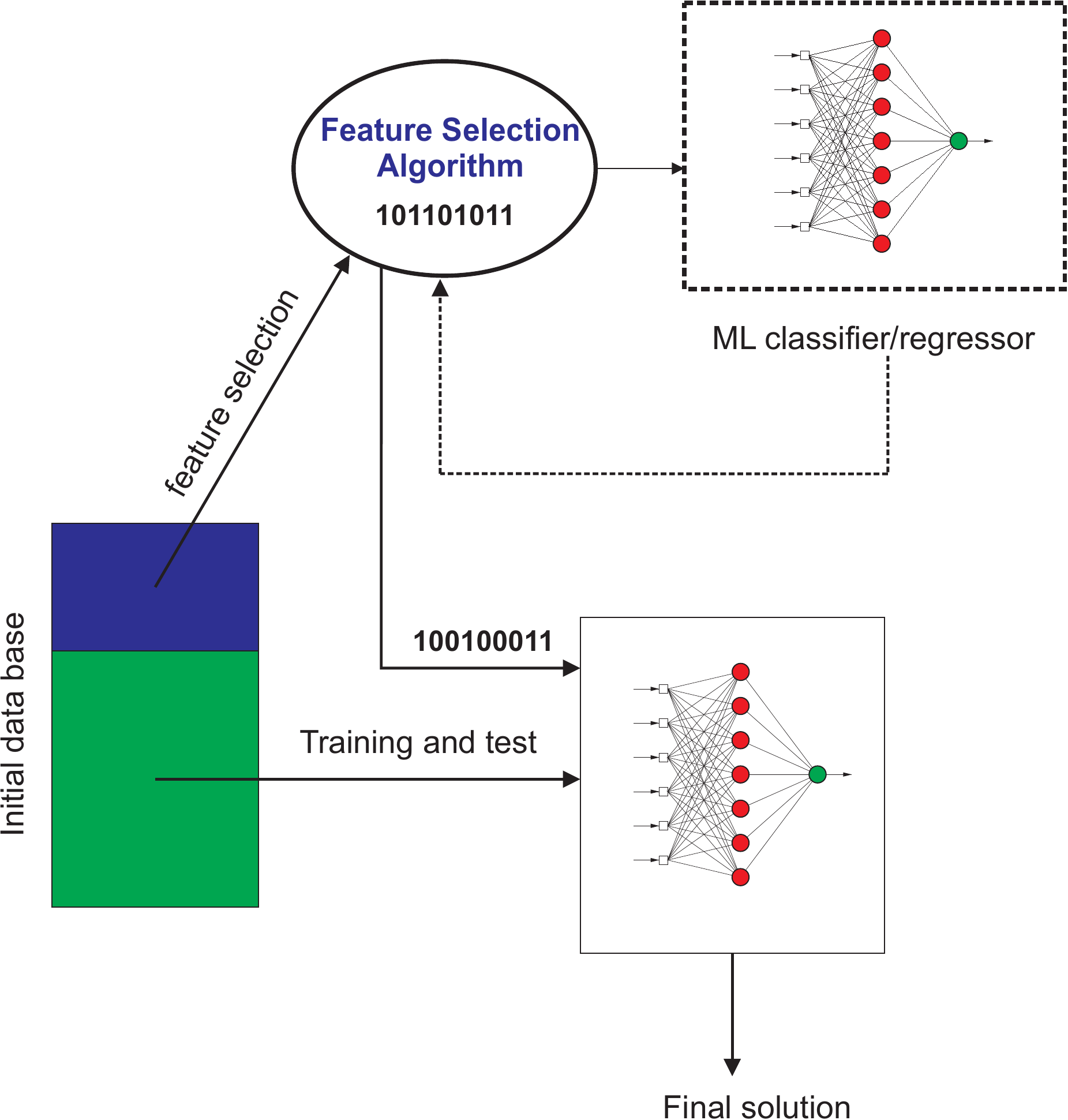}
    \caption{Wrapper}
    \label{fig:wrapper}
\end{subfigure}
\begin{subfigure}[b]{0.45\linewidth}
    \includegraphics[width=\textwidth]{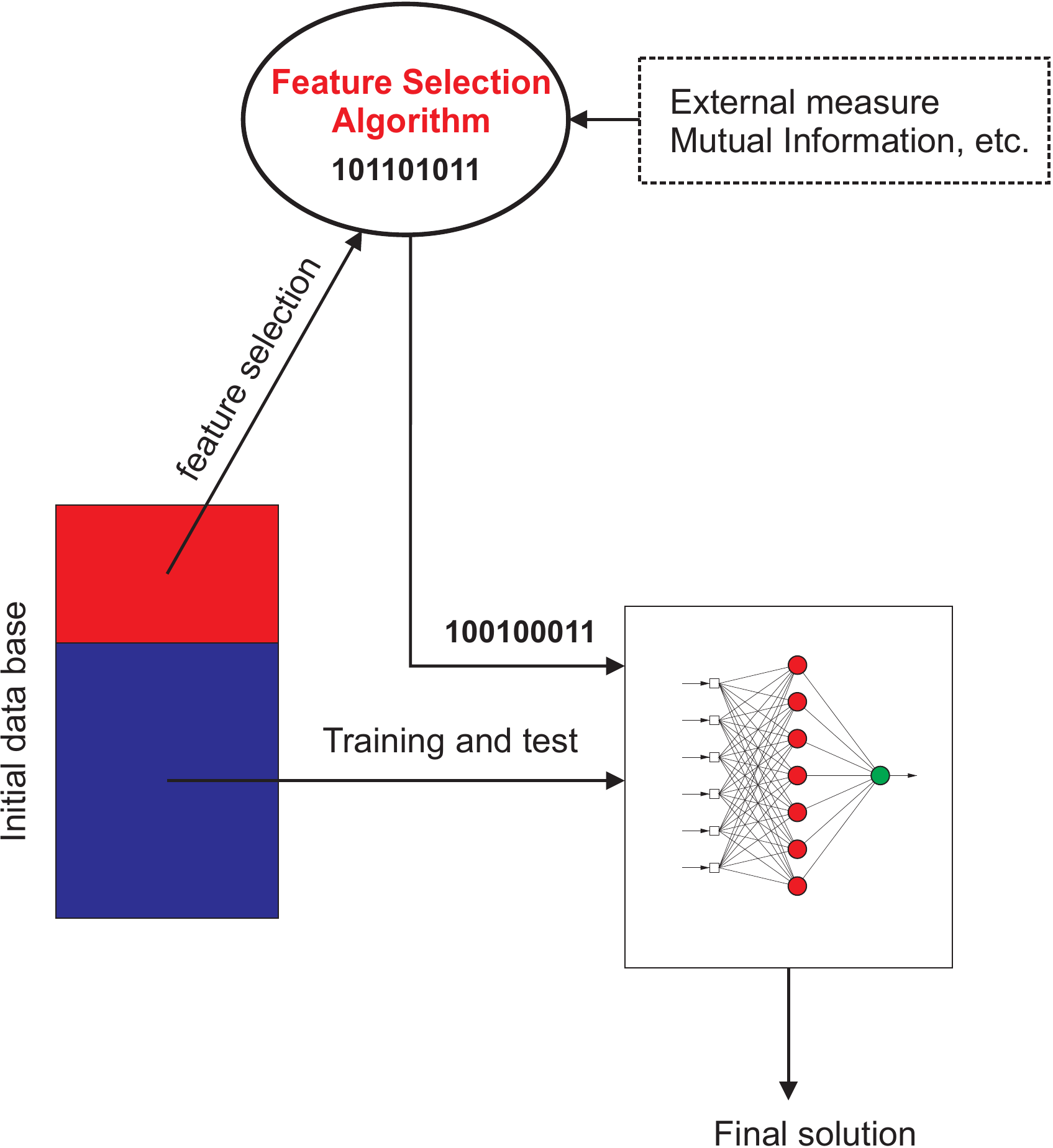}
    \caption{Filter}
    \label{fig:filter}
\end{subfigure}
\caption{\small {(a) Outline of a Wrapper method; (b) Outline of a Filter method.}}
\label{wrapperandfilter}
\end{figure}

There are many algorithms which can be used to solve a FSP. In general, FS algorithms can be classified into three families:

\begin{itemize}
\item The {\em wrapper approach} to the FSP was introduced in \cite{John94}. Wrapper methods
search for a good subset of features using the classifier/regressor itself as part of the evaluating function. Figure \ref{wrapperandfilter} (a) shows the idea behind the wrapper approach: the classifier/regression technique is run on the training dataset with different subsets of features. The one which produces the lowest estimated error in an independent but representative test set is chosen as the final feature set. For further reading, the following classical works can be consulted \cite{Kohavi97,Yang_1}. In the case of the wrapper method, the FSP admits a mathematical definition as follows: The FSP consists of finding the optimum $n$-column vector $\boldsymbol{\sigma}$, where $\sigma_i\in\{1,0\}$, that defines the subset of selected features, which is found as

\begin{equation}
\boldsymbol{\sigma}^{o}=
\argmin_{\boldsymbol{\sigma},\boldsymbol{\alpha}}\left( \int V(y,f({\bf
x}\ast{\boldsymbol \sigma},\boldsymbol{\alpha})) dP({\bf x},y)\right), \label{formulacion}
\end{equation}
where $V(\cdot,\cdot)$ is a loss functional, $P({\bf x},y)$ is the unknown probability function the data was sampled from and we have
defined ${\bf x} \ast \boldsymbol{\sigma}=(x_1\sigma_1, \ldots, x_n\sigma_n)$. The function $y=f({\bf x},\boldsymbol{\alpha})$ is
the classification/regression engine that is evaluated for each subset selection, $\boldsymbol{\sigma}$, and for each set of its
hyper-parameters, $\boldsymbol{\alpha}$.

\item In the {\em filter approach} to the FSP, the feature selection is performed based on the data alone, ignoring the classifier algorithm.
An external measure calculated from the data must be defined to select a subset of features. After the search, the best
feature subset is evaluated on the data by means of the classifier/regression algorithm. Note that the performance of filter algorithms depends entirely on the measure selected for comparing subsets. Figure \ref{wrapperandfilter} (b) shows an example of how
a filter algorithm works. Filter methods are usually faster than wrapper methods. However, they totally ignore the effect of the
selected feature subset on the performance of the classification/regression algorithm during the search, resulting in usually poorer performance than wrapper approaches. Further analysis and applications of filter methods for feature selection can be found in \cite{torkkola00mutual,torkkola02feature}.

\item Finally, mixed or {\em hybrid approach}: there are different works which have combined both wrapper and filter methodologies to build hybrid approaches. They have shown good performance in specific applications \cite{ferreira2014incremental,huda2014hybrid,solorio2016new}.
\end{itemize}

For both wrapper and filter methods, a binary representation can be used for the FSP, where a 1 in the $i_{th}$ position of the
binary vector means that the feature $i$ is considered within the subset of features, and a 0 in the $j_{th}$ position means that feature $j$ is not considered within the subset. Note that using this notation is equivalent to encode
the problem as the vector ${\boldsymbol \sigma}$ included in Equation \eqref{formulacion}. Note also that there are $2^n$
different subsets (where $n$ is the total number of features), and the problem is to select the best one in terms of a
certain measure, which can be either internal (wrapper methods) or external (filter methods) to the classifier/regressor. Alternative encodings, such as integer vectors, are however also possible, and sometimes even more adequate in some specific applications.

\section{Review of existing literature}\label{sec:LitREv}

This section critically analyzes and discusses the existing literature related to ML in atmospheric EEs. The methodology applied has been the following: We perform a large number of search queries in well-known scientific publication databases, including Google Scholar, Scopus and Web of Science. We systematically introduce a specific set of query strings in order to discover published works related to ML in atmospheric EE. We have used the term ML together with \textit{extreme atmospheric events}, plus \textit{extreme rainfall}, \textit{flood prediction}, \textit{heatwaves prediction}, \textit{extreme temperature prediction}, \textit{droughts prediction}, \textit{convective systems}, \textit{tropical cyclones prediction}, \textit{hail and hailstorms}, \textit{extreme wind gusts} or \textit{low-visibility prediction}, among many other terms linked to atmospheric EE. Once all results were retrieved from the aforementioned databases, we removed duplicates and performed an exhaustive analysis and discussion on a paper by paper basis, towards ascertaining their alignment with the topic under study. This systematic review process gave rise to the review and analysis that we present in the subsequent sections.

\begin{figure}[htpb]
	\resizebox{1.0\columnwidth}{!}{\begin{forest}
			upper style/.style = {draw, rectangle, top color=white, bottom color=black!20},
			lower style/.style = {draw, thin, text width=14em,align=left},
			s sep'+=50pt,
			forked edges,
			where level<=1{%
				upper style
			}{
				lower style,
			},
			where level<=0{%
				parent anchor=children,
				child anchor=parent,
				if={isodd(n_children())}{%
					calign=child edge,
					calign primary child/.process={
						O+nw+n{n children}{(#1+1)/2}
					},
				}{%
					calign=edge midpoint,
				},
			}{
				folder,
				grow'=0,
			},
			[{{\Huge Machine Learning Methods for}\\{\Huge Atmospheric Extreme Event Problems}}, for tree={fill=white,minimum size=2cm}
			[{{\huge Rainfall and Floods}}, bottom color=blue!20
			[{{\huge Conventional}}\\
			{{\huge Regresors}}
			\vspace{3mm}\\
			{\huge \cite{moon2019application,diez2020long,vandal2019intercomparison}}\\
			{\huge \cite{hosseini2020flash}}\\
			]
			[{{\huge SVMs}} \vspace{3mm}\\
			{\huge \cite{diez2020long,vandal2019intercomparison}}\\
			]
			[{{\huge Ensembles}} \vspace{3mm}\\
			{\huge \cite{diez2020long,grazzini2020extreme,hosseini2020flash}}\\
			{\huge \cite{hu2019machine,bui2019flash,choi2018development}}\\
			]
			[{{\huge Clustering}} \vspace{3mm}\\
			{\huge \cite{diez2020long,grazzini2020extreme}}\\
			]
			[{{\huge ANNs}} \vspace{3mm}\\
			{\huge \cite{diez2020long,schlef2019atmospheric,shi2020enabling}}\\
			{\huge \cite{vandal2019intercomparison,yeditha2020forecasting}}\\
			]
			[{{\huge DL}} \vspace{3mm}\\
			{\huge \cite{moishin2021designing,xie2021improving}}
			]
			[{{\huge Complex}}\\{{\huge Networks}} \vspace{3mm}\\
			{\huge \cite{boers2019complex}}\\
			]
			] 
			[{{\huge Heatwaves and}}\\{{\huge Extreme Temperatures}}, bottom color=red!20
			[{{\huge Conventional}}\\
			{{\huge Regresors}}
			\vspace{3mm}\\
			{\huge \cite{park2018defining,ahmed2020multi}}\\
			]
			[{{\huge SVMs}} \vspace{3mm}\\
			{\huge \cite{paniagua2011prediction,ahmed2020multi,Oettli2022}}\\
			]
			[{{\huge Ensembles}} \vspace{3mm}\\
			{\huge \cite{peng2020prediction,Oettli2022}}\\
			]
			[{{\huge ANNs}} \vspace{3mm}\\
			{\huge \cite{abdel1995modeling,pasini2017attribution,de2009artificial}}\\
			{\huge \cite{ahmed2020multi,peng2020prediction}}
			]
			[{{\huge DL}} \vspace{3mm}\\
			{\huge \cite{chattopadhyay2020analog}}
			]
			] 
			[{{\huge Droughts}}, bottom color=green!20
			[{{\huge Conventional}}\\
			{{\huge Regresors}}\vspace{3mm}\\
			{\huge \cite{khan2020prediction,rahmati2020machine}}\\
			]
			[{{\huge SVMs}} \vspace{3mm}\\
			{\huge \cite{khan2020prediction,rahmati2020machine,feng2019machine}}\\
			{\huge \cite{belayneh2013drought,belayneh2016short,belayneh2014long}}\\
			{\huge \cite{belayneh2016coupling,dikshit2020temporal,richman20182015}}\\
			{\huge \cite{li2021robust,yaseen2020prediction,tufaner2020estimation}}\\
			{\huge
				\cite{roodposhti2017drought}}
			]
			[{{\huge Ensembles}} \vspace{3mm}\\
			{\huge \cite{sutanto2019moving,rhee2017meteorological,park2016drought}}\\
			{\huge \cite{rahmati2020machine,feng2019machine,zhang2019meteorological}}\\
			{\huge \cite{mokhtar2021estimation,li2021robust,tufaner2020estimation}}\\
			{\huge \cite{prodhan2022projection}}\\
			]
			[{{\huge ANN}} \vspace{3mm}\\
			{\huge \cite{khan2020prediction,feng2019machine,belayneh2013drought}}\\
			{\huge \cite{belayneh2016short,belayneh2014long,belayneh2016coupling}}\\
			{\huge \cite{aghelpour2020theoretical,deo2015application,dikshit2020temporal}}\\
			{\huge \cite{li2021robust,tufaner2020estimation}}\\
			]
			[{{\huge DL}} \vspace{3mm}\\
			{\huge \cite{mokhtar2021estimation,adikari2021evaluation}}\\
			]
			] 
			[{{\huge Severe weather}}, bottom color=yellow!20
			[{{\huge Conventional}}\\
			{{\huge Regresors}} \vspace{3mm}\\
			{\huge \cite{jergensen2020classifying,flora2021using,baki2021determining}}\\
			{\huge \cite{kar2021tropical,nethery2020integrated,wendler2021modeling}}\\
			{\huge \cite{lopez2007short,sallis2011machine,lagerquist2017machine}}\\
			]
			[{{\huge SVM}} \vspace{3mm}\\
			{\huge \cite{jergensen2020classifying,tan2021western,kar2021tropical}}\\
			{\huge \cite{tebbi2016artificial}}
			]
			[{{\huge Ensembles}} \vspace{3mm}\\
			{\huge \cite{jergensen2020classifying,hill2020forecasting,mcgovern2017using}}\\
			{\huge \cite{flora2021using,tan2021western,sun2021machine}}\\
			{\huge \cite{kar2021tropical,kim2021decision,zhu2021evaluating}}\\
			{\huge \cite{ngo2021novel,zhang2021impact,gagne2015day}}\\
			{\huge \cite{gagne2017storm,czernecki2019application,yao2020application}}\\
			{\huge \cite{burke2020calibration,sallis2011machine,shanmuganathan2014data}}\\
			{\huge \cite{lagerquist2017machine,wang2020probabilistic,schulz2021machine}}\\
			{\huge \cite{arul2022machine}}
			]
			[{{\huge ANN}} \vspace{3mm}\\
			{\huge \cite{guijo2020prediction,guijo2020ordinal,sallis2011machine}}\\
			{\huge \cite{shanmuganathan2014data,lagerquist2017machine,spassiani2021application}}\\
			]
			[{{\huge DL}} \vspace{3mm}\\
			{\huge \cite{zhou2019forecasting,asthana2021atlantic,pullman2019applying}}\\
			{\huge \cite{gagne2019interpretable,wang2020probabilistic}}\\
			]
			] 
			[{{\huge Fog and extreme}\\{\huge low-visibility}}, bottom color=magenta!20
			[{{\huge Conventional}}\\
			{{\huge Regresors}} \vspace{3mm}\\
			{\huge \cite{Boneh15,bartokova2015fog,Guijo18}}\\
			{\huge \cite{bari2020machine,CASTILLOBOTON2022106157}}\\
			]
			[{{\huge SVM}} \vspace{3mm}\\
			{\huge \cite{Cornejo17,cornejo2020persistence,CASTILLOBOTON2022106157}}\\
			]
			[{{\huge Ensembles}} \vspace{3mm}\\
			{\huge \cite{dietz2019forecasting,bartokova2015fog,bari2020machine}}\\
			{\huge \cite{li2020diagnostic,yu2021application,CASTILLOBOTON2022106157}}\\
			]
			[{{\huge ANN}} \vspace{3mm}\\
			{\huge \cite{Fabbian07,Colabone15,Cornejo17}}\\
			{\huge \cite{Duran18,bari2020machine,cornejo2021statistical}}\\
			{\huge \cite{CASTILLOBOTON2022106157}}\\
			]
			[{{\huge DL}} \vspace{3mm}\\
			{\huge \cite{zhu2017application,miao2020application}}\\
			]
			[{{\huge Fuzzy Logic}} \vspace{3mm}\\
			{\huge \cite{Miao12}}\\
			]
			] 
			] 
	\end{forest}}   
	\centering
	\caption{Summary of recent works dealing with ML, DL and related techniques in atmospheric extreme event problems.}
	\label{fig:taxonomy}
\end{figure}
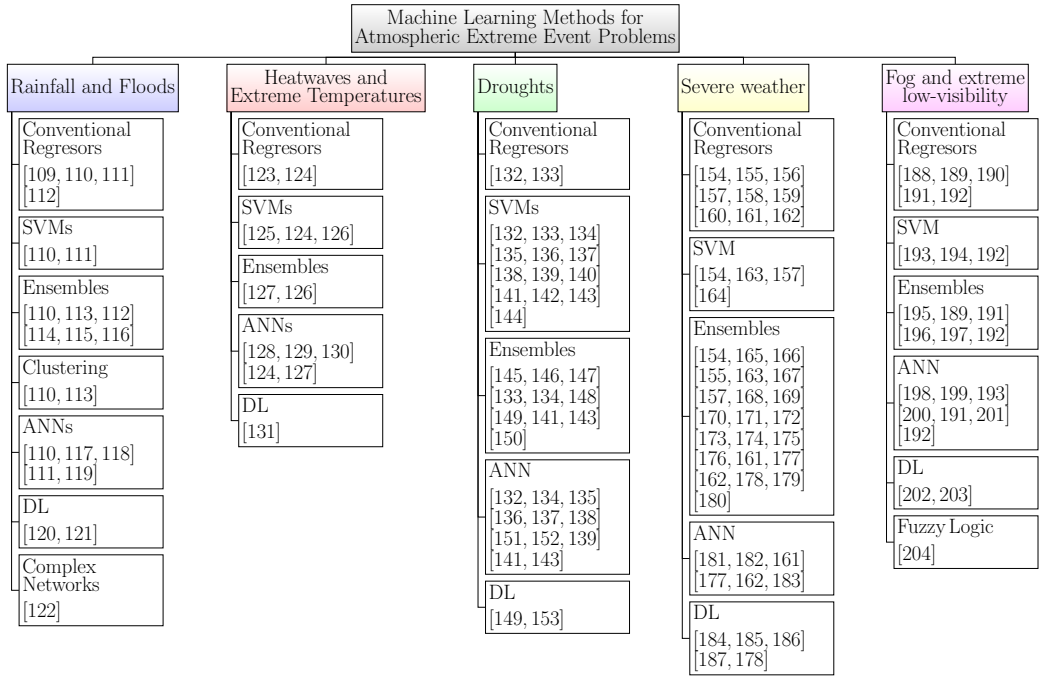

Figure~\ref{fig:taxonomy} summarizes the hierarchical categorization of the state-of-the-art methods for atmospheric EEs problems. We classify the works according to the atmospheric event they predict, and then, using the type of ML methods they involve. Some works are included in several boxes since they apply several ML methods in EEs prediction problems.

\subsection{Extreme rainfall and floods}
Destructive extreme precipitation events and flooding episodes are a real threat to human settlements in different parts of the world \cite{madsen2014review,berghuijs2017recent}. Extensive research on the monitoring, prediction and analysis of these events have been carried out in the literature. We analyze here those works dealing with ML techniques. Note that a first review on ML for floods prediction can be found in \cite{w10111536}, where the state of the art in this topic can be found, up to 2018.
In \cite{moon2019application} a ML-based early warning system for short-term heavy rainfall is proposed for Korea. The system is formulated as a binary classification problem, where a logistic regression has been implemented over predictive variables from meteorological data obtained from automatic weather stations, which have been previously preprocessed by applying a principal component analysis algorithm. A comparison against early warning systems formed by alternative classifiers is carried out. An important amount of meteorological variables measured at different locations feed the classifiers in real time, in order to improve the performance of the classification output.
In \cite{diez2020long} a number of ML methods (SVM, k-nearest neighbours, RF, k-means clustering and neural networks) are applied to a problem of long-term rainfall prediction, using the atmospheric synoptic patterns as predictive variables. Neural networks are reported as the most accurate method, but surprisingly, the work reports the generalized linear method with gamma-distributed errors as the best method to predict the extreme of the series, improving the performance of the ML approaches. Note that supervised and non-supervised methods (k-means) are tested together, and depending on the method, a classification or regression problem is considered, which is an unusual procedure in the application of ML techniques. Results considering as ground truth rain gauges measurements from Tenerife (Canary Islands, Spain), are discussed.
In \cite{schlef2019atmospheric} a self-organized map is used to obtained clusters of synoptic situations leading to extreme floods across USA. Then the flood characteristics of each synoptic situation are analyzed, identifying four primary categories of circulation patterns with different flood potential hazard. This methodology also allows identifying regions where extreme floods occur outside the normal flood season, and other regions where multiple extreme flood events occur within a single year, mainly due to tropical cyclones.


In \cite{shi2020enabling} Convolutional Neural Networks (CNN) are used to carry out a smart dynamical downscaling of extreme convective precipitation from Global Climate Models (GCM). This work shows that when trained with data for three subtropical/tropical regions, CNNs are able to retain between 92\% and 98\% of extreme precipitation events.
In \cite{vandal2019intercomparison}, a problem of extreme precipitation statistical downscaling of GCM is tackled with ML algorithms. Five ML methods are compared in this task: Ordinary Least Squares, Elastic-Net, and Support Vector Machine, Sparse Structure Learning (MSSL) and Autoencoder Neural Networks. Experiments with data from Northeastern United States suggest that the direct application of ML techniques does not improve the results of simpler statistical-based methods in the downscaling of extreme precipitation events.

In \cite{grazzini2020extreme} the classification of precipitation extreme events in northern-central Italy is carried out by means of K-means clustering and RF algorithm. The study reports the importance of integrated water vapour transport variable in the correct detection of extreme precipitation events in this region. This work has been complemented with a second study for the same zone, where the authors investigate the connection between precipitation extremes and Rossby wave packets \cite{grazzini2021extreme}.
In \cite{jahangir2019spatial} an ANN algorithm is applied for the prediction of discharge values and spatial modeling of floods in Kan River Basin, Iran. Similarly, in \cite{yeditha2020forecasting} different ML models (mainly neural networks) with a previous data treatment by wavelets are applied to forecast extreme precipitation from satellite measurements. The proposed approach has been tested in the prediction of floods in Vamsadhara river basin, India.

In \cite{hosseini2020flash} a problem of flash flood forecasting with ML algorithms is tackled. The paper analyzes an ensemble of boosted generalized linear models random forest, and Bayesian generalized linear models algorithms. A pre-processing step for reducing the number of input variables with a Simulated Annealing algorithm is carried out. These approaches are tested in the prediction of flash floods in the North of Iran.
In \cite{hu2019machine} a Gradient Boosting Tree algorithm is applied to perform projections of precipitation intensity over short durations events, using outputs from GCMs. The algorithm performance has been tested in observational data (25 years of data) across USA.
In \cite{bui2019flash} an approach for flash flood susceptibility modeling is proposed. The algorithm combines tree-based ensemble with a pre-processing step of feature selection using a fuzzy-rule method and a Genetic Algorithm. These approaches have been combined with different tree-based ensembles such as LogitBoost, Bagging and AdaBoost algorithms. The performance of the systems was tested in data from Lao Cai Province (Northeast Vietnam).
In \cite{choi2018development} different ML classification techniques such as decision trees, bagging, random forests or boosting have been applied to the prediction of heavy rain damages at Seoul (South Korea). The work uses data on the occurrence of heavy rain damages in the city from 1994 to 2015, obtaining accurate results specially with the boosting technique.
Deep learning approaches have been recently applied to floods prediction. In \cite{moishin2021designing} a CNN with LSTM Network has been introduced to forecast the future occurrence of flood events. The performance of this deep learning approach has been tested in 9 different rainfall datasets of floods occurred in Fiji. In \cite{xie2021improving} a problem of short-term intensive rainfall prediction was tackled with deep learning approaches. ECMWF forecast data and ground observation station data were taken into account, and K-means, generative adversarial nets and deep belief networks were applied to obtain the prediction as a classification model. Experiments in data from the Fujian Province (southeastern China) in the period 2015-2018, showed a good performance of the proposed prediction approaches, improving the results of LSTM and Stacked Sparse Autoencoder networks.

Finally, in close connection with ML approaches, Complex Networks (CN) have also been used to analyze problems of extreme precipitation. In \cite{boers2019complex} the teleconnections of extreme events over the world are studied, using the CN paradigm  over high-resolution satellite data. The CN methodology confirms Rossby waves as the physical mechanism behind global teleconnection patterns in extreme precipitation events.

\subsubsection{Analysis}
As final note on the application of ML models to EEs related to rainfall and floods, we have found ML approaches in very different applications, including short-term and long-term detection and prediction problems, tackled with different ML frameworks (classification and regression) and considering very different prediction (or detection) time horizons.
It is also remarkable the different ways in which many of these approaches introduce the physics of the problem within their approaches. In some cases, mainly in short-term prediction problems, the revised works consider real time meteorological variables to feed ML algorithms, such as in \cite{moon2019application}. In other cases, the ML extract information from synoptic patterns, mainly in problems of long-term rainfall and floods prediction (\cite{diez2020long,schlef2019atmospheric}). In other cases, the output of GCM are treated with ML approaches in order to obtain improvements on the prediction of heavy precipitation events \cite{shi2020enabling,vandal2019intercomparison,hu2019machine}. Other ML approaches rely on specific variables from reanalysis data, but including in the studies variables with physical sense, such as sensitive to flow conditions and other representative of thermodynamic conditions for extreme precipitation events modelling, such as \cite{grazzini2020extreme}. A final group of works have been revised which only rely on measurements or set of data, without any specific consideration of the physics of the problem, specially when DL have been applied (\cite{moishin2021designing,xie2021improving}), but also with shallow ML approaches (\cite{choi2018development}). In these last cases, the works analyzed seem to focus on the ability of ML approaches to extract information and obtaining accurate predictions, evaluated from different metrics, and compared against other ML approaches, with very few references to the physical processes causing the EE. The work in \cite{boers2019complex} analyzes extreme precipitation events from CN paradigm, generating networks which take into account the physics of the problem and the relationship among different variables involved in the problem, including the analysis of teleconnections. 

\subsection{Heatwaves and extreme temperatures}\label{sec:heatwaves}
Extreme temperatures \cite{barriopedro2011hot,pfleiderer2018quantification}, heatwaves \cite{chapman2019warming} and, in the last decades, mega heatwaves \cite{bador2017future,sanchez2018june} are among the extreme atmospheric events potentially most dangerous for people, specially the elderly \cite{diaz2002heat,diaz2002effects} and with deep societal impact. The detection and attribution of heatwaves and extreme temperatures is therefore a hot topic in atmospheric EEs research \cite{wang2017detection}, including the study of natural causes such as circulation patterns \cite{shi2018trends} or anthropogenic contribution \cite{zwiers2011anthropogenic}. ML methods have been applied to study these and other aspects of extreme temperatures and heatwaves \cite{cifuentes2020air}.

\subsubsection{Heatwaves}
In \cite{pasini2017attribution} neural computation is used in a problem of attribution of heatwaves. The study considers the last 160 years, where the attribution to anthropogenic forcings is obtained for the last 50 years, whereas in the period 1910-1975 the main driver is the solar irradiation. The study also clarifies the role of aerosols and the Atlantic Multidecadal Oscillation in decadal temperature variability.
In \cite{park2018defining} Multivariate Adaptive Regression Splines are used to set appropriate heatwave thresholds, in order to improve early warning systems for these events. The work uses daily data of emergency patients diagnosed with heatstroke and also information on 19 meteorological variables obtained for the years 2011 to 2016. The results obtained shown that the combination of heat illness data and average daytime temperature (from noon to 6 PM) can be used as an alternative threshold for heatwaves characterization. Finally, in \cite{chattopadhyay2020analog} a hybrid approach combining the Analog prediction method (search of analogue synoptic situations in the past) with deep neural networks (capsule neural networks, CapsNets) is proposed to predict heatwaves and cold spells. The proposed CapsNets outperformed other deep approaches such as CNN and alternative prediction algorithms such as logistic regression techniques.

\subsubsection{Extreme temperatures}
One of the first approaches in the application of ML techniques for extreme temperature prediction was \cite{abdel1995modeling}, where different artificial neural networks models are applied to a problem of daily maximum temperature prediction in Dhahran, Saudi Arabia. In this case daily data for 18 weather parameters are considered as input variables, to predict the maximum temperature on a given day, with different prediction time horizons up to 3 days in advance.
In \cite{paniagua2011prediction} a SVR algorithm is used to forecast daily maximum air temperature with a 24h prediction time horizon. The prediction system relies on a number of input variables such as air temperature, precipitation, relative humidity and air pressure. It also considers the synoptic situation of the day in order to improve its results. The performance of the SVR algorithm has been successfully evaluated with data from a number of European measurements stations.
In \cite{de2009artificial} the prediction of the maximum (and minimum) air temperature in the summer monsoon season is carried out by using a multi-layer MLP perceptron neural network. The mean temperature of previous months in the period of analysis is considered as inputs for the system. Data from the Indian Institute of Tropical Meteorology belonging to the years 1901-2003 are considered.
In \cite{ahmed2020multi} different ML approaches such as MLP, SVM and Relevance Vector Machine (RVM) or K-Nearest Neighbour (KNN), are proposed to develop multi-model ensembles from global climate models. The objective is to obtain annual prediction of monsoon and winter precipitation, maximum temperature and minimum temperature over Pakistan. The results obtained have shown that KNN and RVM-based multi-method ensembles show better skills than those developed with MLP and SVM.
In \cite{peng2020prediction} a MLP and and a natural gradient boosting algorithm (NGBoost), are applied to improve the prediction skills of the 2-m maximum air temperature, with prediction time horizon with lead times from 1 to 35 days. The ML prediction approaches have shown better results than the ensemble model output statistics (EMOS) method (which was selected as the benchmark for comparison) in 90\% of the cases analyzed.
In \cite{Oettli2022} a number of ML algorithms such as neural networks, SVMs, RF, Gradient Boosting or regression trees have been applied to the prediction of surface air temperature two months in advance, with input data two months in advance from SINTEX-F2, a dynamical prediction system. The dynamical prediction system includes the physics of the problem, while the ML algorithms improves the results by a statistical downscaling. The performance of these approaches has been tested in Tokio (Japan), obtaining excellent prediction results. 

\subsubsection{Analysis}

The works revised in this subsection reveal that there are not many works dealing with heatwaves prediction using ML approaches. Only two specific works on application of ML techniques to heatwaves estimation have been found in the recent literature. In the first case \cite{park2018defining}, the work uses data from meteorological variables and emergency patients in order to obtain characterization of heatwaves. In the second approach \cite{chattopadhyay2020analog} ML algorithms (DL networks in this case) are merged with Analog method which introduces the physics of the problem in order to predict heatwaves. There are many more works on ML algorithms for extreme temperatures prediction problems. Artificial neural networks and statistical ML approaches are the main algorithms applied in the literature to tackle these problems. It is interesting to see how in these works, the inclusion of the physics is not as relevant than in the works dealing with ML algorithms for rainfall and flood prediction. The reason for this is that air temperature is in general a variable easier to be predicted than rainfall, in which the inclusion of the atmospheric state and dynamics is key to obtain good results. Synoptic situations (considered in \cite{paniagua2011prediction}) seems to improve the results of ML algorithms in the prediction of extreme temperatures. In the rest of articles revised, the prediction is based on existing registers of previous temperatures. The application of ML approaches produces good results in this case in weekly or monthly temperature predictions, where the variation of the extreme temperatures is small.

\subsection{Droughts}

Droughts are extreme events, stochastic in nature, with a deep impact on society, specifically on water supplies, agriculture, hydroelectric power production, and associated with forest fires and even forced migrations \cite{spinoni2019new,garcia2019european}. Drought early warning systems provide important information about predicted drought hazards. In many cases, these systems rely on ML algorithms.

In \cite{sutanto2019moving} a RF algorithm is used to forecast drought impacts, by relating forecasted hydro-meteorological drought indices to previously reported drought impacts. The proposed model based on ML is able to forecast drought impacts with prediction time horizons of some months ahead.
In \cite{khan2020prediction} different ML classification techniques are applied to develop drought prediction models over Pakistan. They include SVM, MLP and KNN algorithms. Meteorological variables from reanalysis are considered as inputs, whereas the objective variable considers three categories of droughts: moderate, severe, and extreme in different cropping seasons. These classes were estimated using Standardized Precipitation Evaporation Index (SPEI; \cite{Vicente-Serrano2010}), in order to train and test the proposed ML classifiers.
In \cite{rhee2017meteorological} a problem of high-resolution spatial drought forecasting is tackled in Korea from remote sensing and climate indices inputs. The performance of different regression trees algorithms, RF and Extremely randomized trees have been compared.
In \cite{park2016drought} different ML algorithms such as RF, boosted regression trees, and Cubist are applied to model meteorological and agricultural droughts from 16 inputs drought factors obtained from satellite measurements. The SPI and crop data are used as objective variables to model the droughts. RF has been reported as the best performance algorithm in data from arid zones of the United States. 
In \cite{rahmati2020machine} drought hazard is tackled with different ML models: classification and regression trees (CART), boosted regression trees (BRT), RF, multivariate adaptive regression splines (MARS), flexible discriminant analysis (FDA) and SVM. Some Hydro-environmental datasets are used to calculate the relative departure of soil moisture (RDSM), and this index is used as objective variable, whereas the inputs are eight environmental factors as potential predictors of drought. Experiments in south-east part of Queensland, Australia, are carried out to evaluate the performance of the different ML methods proposed.
In \cite{feng2019machine} three ML algorithms (RF, SVM and MLPs) are used to evaluate whether remotely-sensed drought factors (satellite measurements) are good estimators for drought events prediction in south-eastern Australia. RF is again the ML regression technique which best results obtains in this problem, outperforming SVM and MLPs in this task.
In \cite{belayneh2013drought} short-term drought prediction in the Awash River Basin (Ethiopia) is considered, by means of SPI prediction. Three ML methods are evaluated for this problem, MLP, SVM and MLP with a previous step of wavelets signal decomposition. The coupled wavelet-MLP algorithm showed the best result in SPI prediction with prediction time horizon of 1 month and 3 months. New results and further analysis on the same problem were reported in \cite{belayneh2016short}.
In \cite{belayneh2014long} a long-term drought prediction problem in the Awash river is considered by means of MLPs and SVMs, enhanced with wavelets transforms. The SPI at 12 and 24 months (SPI 12 and SPI 24) are predicted by means of the ML methods. Comparison with ARIMA methods for time series prediction shows a better performance of the ML techniques.
The same data from Awash River Basin are used in \cite{belayneh2016coupling} to test advanced versions of ML algorithms in the same problem of drought prediction. Coupled versions of ML algorithms with wavelet transforms are considered, such as wavelets transforms with Bootstrap and Boosting ensembles together with MLP and SVR models. These coupled models show a better performance than the MLP and SVR algorithms on their own.
In \cite{roodposhti2017drought} a problem of drought sensitivity mapping based on SPI index and enhanced vegetation index (EVI) is tackled, by using one-class SVMs. Data from both synoptic stations and satellite data are combined in this study in the Iranian province of Kermanshah.  
In \cite{deo2015application} the performance of the ELM algorithm is evaluated in a problem of Effective Drought Index prediction in eastern Australia. Predictive variables composed of meteorological variables and climate indices are considered. The ELM approach outperformed the results of different neural networks models.
In \cite{aghelpour2020theoretical} different ML approaches are evaluated in a problem of forecasting the precipitation joint deficit index (JDI) and the multivariate standardized precipitation index (MSPI), both of them related to severe droughts. Different ML methods are considered, such as group method of data handling (GMDH), generalized regression neural network (GRNN), least squared support vector machine (LSSVM), adaptive neuro-fuzzy inference system (ANFIS) and ANFIS optimized with meta-heuristics algorithms. Experiments in data from 10 measuring stations in Iran are considered. The GMDH method is reported as the most accurate algorithm.
In \cite{zhang2019meteorological} artificial neural networks and eXtreme Gradient Boost (XGBoost) algorithms with feature selection by means of a cross-correlation function and a distributed lag nonlinear model (DLNM) are considered in a problem of drought prediction. Data from 32 stations during 1961 to 2016 in the Shaanxi province, China, are used. The results show that the XGBoost approach outperforms neural networks and the DLNM works better than the cross-correlation function in the selection of the best features for this prediction problem.
In \cite{mokhtar2021estimation} four ML methods (RF, the Extreme Gradient Boost (XGB),  Convolutional neural networks (CNN) and the Long-term short memory (LSTM)) are considered in a problem of SPEI estimation in the Qinghai-Tibet Plateau. Meteorological variables and climate indices are considered as predictive variables.
In \cite{dikshit2020temporal} MLP and SVR algorithms are tested in a problem of drought prediction in New South Wales, Australia. SPEI index at 1, 3, 6 and 12 months are used as objective value. The results obtained suggest that the MLP outperforms SVM. The results also discard that sea temperature and climate indices had a real impact on the droughts at New South Wales.
In \cite{richman20182015} a Feature Selection Problem is considered for attribution of the Cape City drought 2015-2017 with ML algorithms. Wrapper algorithms for FSP are considered, in which the SVM has been used as classification algorithm, and different evolutionary algorithms look for the best set of features (drought drivers) for predicting the cool season precipitation in the years of the drought.
In \cite{li2021robust} the role of antecedent SST fluctuation pattern (ASFP) as drought driver is analyzed by using ML techniques such as SVR, RF and ELM. The SPEI is used as objective to be predicted at different river basins such as Colorado, Danube, Orange, and Pearl river. The obtained results showed that the ASFP-ELM model can effectively predict space-time evolution of drought events outperforming the rest of the ML algorithms considered.
In \cite{prodhan2022projection} RF and Gradient Boosting Machine algorithms are applied to characterize future drought metrics, and its impact on crops. The magnitude, intensity, and duration of future droughts are characterized by means of the SPEI drought index using CMIP6 (Coupled Model Inter-comparison Phase-6) climate models data. Experimental results on Southern Asia, including countries such as Afghanistan, Pakistan and India are analyzed.

In close connection with droughts forecasting, evaporation prediction has been tackled in some cases.
For instance, recently \cite{yaseen2020prediction} evaluates ML approaches for evaporation prediction in arid regions of Iraq. Four different ML models are considered including classification trees, a cascade correlation neural network, a gene expression programming (GEP), and a SVM algorithm.
Another recent work dealing with alternative prediction problems related to drought forecasting is, \cite{tufaner2020estimation} where the Palmer Drought Severity Index (PDSI) is predicted by using different ML algorithms. SVM, MLP and decision trees have been applied to this problem, and their results compared to a Linear Regression algorithm used as baseline technique. Results in a problem of PDSI prediction in Anatolia (Turkey), has shown that the MLP obtains the best results.
Finally, \cite{adikari2021evaluation} evaluates the performance of three different ML algorithms (Convolutional Neural Networks (CNN), Long-Short Term Memory network (LSTM), and Wavelet decomposition functions combined with the Adaptive Neuro-Fuzzy Inference System (WANFIS)) in two different problems of flood and drought forecasting. The results obtained reveal that CNNs is the best compared approach for flood forecast and WANFIS outperforms the other two algorithms in drought forecasting.

\subsubsection{Analysis}

The review of articles about ML techniques to drought and related problems has shown a large number of ML algorithms applied to drought prediction and analysis. Ensemble methods such as RF seems to be strong approaches for prediction problems related to drought, thought other algorithms such as neural networks, statistical learning approaches and DL algorithms have also been successfully applied to different drought prediction cases. The inclusion of the physics is, in the majority of cases, treated by means of considering climate indices among the predictive (input) variables of the problems, though some approaches such as \cite{dikshit2020temporal} have discarded that climate indices improve as predictive variables improve the performance of ML algorithms in specific problems of drought prediction. In general, processes related to atmospheric dynamics seems to dominate this phenomenon, so the inclusion of climate indices as inputs for ML algorithms seems a reasonable election in order to capture the physics of the problem. Regarding the objective variables for defining the problem, the majority of problems analyzed used precipitation indices such as SPI or SPEI, as drought indicators.

\subsection{Severe weather}
EEs related to severe weather have also been studied and analyzed with ML methods in the last years. We have divided this subsection into different parts, ML methods in convective systems studies, tropical cyclones, hailstorms and extreme wind and gusts. 

\subsubsection{Convective systems}
There are different works focused on the study of convective clouds and systems formation and related events with ML approaches \cite{xiu2016identification}.
In \cite{tebbi2016artificial} a problem of convective cloud classification by means of the combination of ANN and SVM, using high resolution satellite images in northern Algeria is tackled. The proposed system works in two steps. First, the system detects rainy areas in cloud systems, and second, it delineates convective cells from stratiform ones.
In \cite{sahoo2019prediction} a problem of storm surge and coastal floods prediction with artificial neural networks is tackled. The work is focused on Odisha state (India), trying to simulate the effects in tide caused by super cyclone of 1999. Comparison with the ADCIRC prediction model \cite{luettich1992adcirc} shows that the ML-based model is able to obtain significant results in the prediction of storm surge and associated flood of Odisha event. 
In \cite{guijo2020prediction} a problem of classification of convective situations over Madrid-Barajas airport is tackled, with neuro-evolutionary techniques (neural networks trained with evolutionary computation techniques). The problem is considered as a multi-class classification problem, highly imbalanced (there are much less convective situation than clear days). However, the neuro-evolutionary approaches are able to obtain an accurate performance in the identification of days with convective clouds formation in Madrid airport. 
A similar problem is tackled in \cite{guijo2020ordinal} by considering ordinal regression techniques instead of classification.
Another study is presented in \cite{jergensen2020classifying}, where a problem of thunder storms classification is tackled with different ML approaches, such as logistic regression algorithms, RF, gradient-boosted forests and SVMs. The problem has been formulated as a multi-class classification problem, in which the gradient-boosted forest algorithm obtained the best classification results.
In \cite{hill2020forecasting} a RF algorithm is evaluated in problems related to convective systems. The study includes different EEs from convective systems such as the presence of tornadoes, large hail (over 1 inch) or induced wind gusts over 58 mph. A large number of predictive variables are considered in this study, including different atmospheric fields such as 10-m winds, surface temperature and specific humidity, precipitable water, accumulated precipitation, wind shear from the surface at different pressure levels or mean sea level pressure, among other. The RF algorithm was able to obtain relationships between predictive atmospheric fields and observations according to the community's physical understandings about severe weather forecasting.
Dealing with a similar idea, \cite{mcgovern2017using} evaluates the performance of RF and Gradient Boosted Regression Trees in a problem of prediction skill for multiple types of high-impact events related to convective systems, such as severe wind, hail or heavy rain, with discussion on the impact of this severe weather in renewable energy or aviation turbulence.
In \cite{zhou2019forecasting} a CNN is introduced for severe convective weather prediction, including heavy rain, hail, convective gusts, and thunderstorms. The predictive variables are obtained from a numerical weather model (Global Forecasting System), and the performance of the CNN is compared to that of traditional methods and human expert evaluation of the data. The results showed that the CNN obtained results which improved the performance of previous algorithms and human expert results, but with some flaws such as a too many false alarms in predicting hail and convective gusts.
Finally, in \cite{flora2021using} three ML approaches RF, gradient-boosted trees, and logistic regression algorithms have been proposed to predict whether ensemble storm tracks will produce a tornado, severe hail, and/or severe wind report. The paper describes a postprocessing using the ML algorithms of the ensemble output from the National Oceanic and Atmospheric Administration Warn-on-Forecast (WoF) project. The results obtained have shown that the ML-based postprocessing of WoF data improves short-term, storm-scale severe weather probabilistic guidance.

\subsubsection{Tropical cyclones}
Other EEs associated with severe weather are Tropical Cyclones (TC). In addition to their extreme associated gusts, they always come with other severe weather events such as heavy rain, hail, or thunder storms, in many occasions deriving in catastrophic events such as floods, ground slides, etc. \cite{chen2020flood}. There is a very recent comprehensive review on ML approaches in TC forecast \cite{chen2020machine}. That article covers previous works on ML for TC up to 2020.
There have been some works dealing with topics related to ML for TC after that review paper. For example there is some recent work dealing with ML for TC prediction and characterization, such as \cite{baki2021determining} where a Multivariate Adaptive Regression Splines (MARS), has been applied to obtain the optimal values of the WRF meso-scale model parameterizations for TC prediction in the Bay of Bengal.
In \cite{tan2021western} a Gradient Boosting Decision Tree model has been proposed for TC track forecast at Western North Pacific. A comparison with climatology and persistence is carried out to evaluate the performance of the proposed ML technique in this problem.
In \cite{sun2021machine} ensemble methods optimized by ML approaches such as Lasso optimization or Ridge regression are proposed to improve preseason prediction of Atlantic hurricane activity.
In \cite{pillay2021conditions} an analysis of the initialization variables affecting TC formation is carried out. RF algorithms are proposed to analyze the importance of each climate variable considered. The RF models are also used to predict intensification magnitudes of the TC based on the state of the input variables.
In \cite{asthana2021atlantic} a CNN was used to predict Atlantic hurricane activity from reanalysis data. Accurate prediction results are reported, in comparison with alternative state of the art models.
In \cite{kar2021tropical}, a different ML algorithm has been applied to a problem of cloud intensity classification in TC over the Bay of Bengal and the Arabian sea. Five ML classifiers have been proposed for this problem: Na\"ive Bayes, SVM, Logistic Model Tree, Random Tree and RF. The RF algorithm showed the best performance over the rest of tested classifiers for this problem.
In \cite{kim2021decision} a decision-tree algorithm has been proposed for a problem of TC maximum lifetime intensity. The algorithm predicts the probability that a TC reaches a maximum  intensity larger than 70 knots. Accurate results are obtained with classification rates over 90\% in the considered test set. There have been some works dealing with the estimation of the precipitation produced by TC using ML techniques.
In \cite{zhu2021evaluating} a RF method is applied to a problem of prediction of the precipitation associated with TC in Eastern Mexico.
In \cite{ngo2021novel} a hybrid Quantum PSO algorithm and a Credal Decision Tree (CDT) ensemble have been proposed for spatial prediction of the flash floods in TC. Experiments are carried out for north-western mountainous area of Vietnam. Satellite data from Sentinel-1 C-band SAR images are considered in this case to model the objective function. Finally there are some recent works dealing with ML applications for evaluating impacts of TC. 
In \cite{nethery2020integrated} ML algorithms, mainly Bayesian methods, are used to estimate health problems caused by TC.
In \cite{wendler2021modeling} the economic impact of TC is analyzed by means of ML approaches, and in \cite{zhang2021impact} the impact of typhoon Lekima on different chinese forests is evaluated by means of RF over Landsat 8 OLI images.

\subsubsection{Hailstorms}
Hail is an atmospheric EEs which causes important economic problems in many countries, mostly in agriculture and crop losses. Though it is not a frequent EEs (returning periods of severe hailstorms have been set around 20 years, depending on the zone, according to different studies \cite{fraile2003return}) there are some works on prediction and characterization of this EE, including the use of ML techniques in the last years. Note, however, that prediction of hailfalls is a difficult task, due to the local spatial characteristic of this EEs and its short duration, which makes that prediction approaches should be developed separately for specific geographic areas.

One of the first works dealing with a prediction problem of hailfalls is \cite{lopez2007short}, in which the problem is tackled as a binary classification task (hail/no-hail). A logistic regression was then applied, obtaining a probability of Detection of 0.87 with a False Alarm Ratio of 0.18. After this initial work on hailstorms prediction, some more sophisticated ML methods were introduced.
In \cite{gagne2015day} a hybrid approach mixing NWM with ML algorithms is proposed for a problem of hailfalls forecasting. The NWM identifies potential
hail storms and different ML algorithms mainly RF and gradient boosting trees are used to predict the hail occurrence. Observed hailstorms are used to obtain the ground truth values for this problem.
In \cite{gagne2017storm} a storm-based probabilistic hail forecasting is proposed, including a RF algorithm in the system. The prediction starts with an identification and tracking algorithm based on radar grid data and convection allowing model. Different parameters for characterizing the storm are then obtained, and passed to the RF algorithm which has been previously trained with data from observed hailstorms. The RF algorithm uses this  information to predict the probability of a storm producing hail, and also provide the hail size estimation.
In \cite{czernecki2019application} a RF algorithm has been proposed for a problem of large hail prediction. Different predictive variables such as radar reflectivity, EUCLID lightning detection data, and convective indices from the ERA5 reanalysis are considered. The objective variables are obtained from observational data of large hail reports from Poland in the period 2008-2017.
Also dealing with hail prediction using a RF algorithm, \cite{yao2020application} used hail observation data from 41 meteorological stations in Shandong Peninsula, China, in the period 1998-2013 to train the algorithm. Different thermal factors and variables such as lifted index, Showalter stability index, and total index are used as predictive variables of hailfalls in this work.
Another example of the use of RF in hail prediction is \cite{burke2020calibration}, in which different observational datasets were used to train and test the RF approach, such as the Maximum Estimated Size of Hail (MESH), and the Multi-Radar Multi-Sensor (MRMS) product. Finally, Some recent works have applied deep learning approaches to problems of hail prediction.
In \cite{pullman2019applying} a deep learning network has been applied to a problem of hailstorms detection. The GOES satellite imagery and MERRA-2 reanalysis data are used as predictive variables in this case.
In \cite{gagne2019interpretable} a CNN is applied to a problem of predicting the probability of severe hail (larger than 2.5 mm) in the next hour. Data for this study have been obtained from NCAR convection-allowing ensemble in May 2016.

\subsubsection{Extreme winds and gusts}
Extreme Wind Gusts (EWG) are associated with severe weather. They can have catastrophic effects on crops and buildings, and also have impact in renewable energy facilities such as wind farms. A first review of techniques for WG prediction, including NWM and also ML approaches has been presented in \cite{sheridan2018current}.
In \cite{sallis2011machine} several ML algorithms have been applied to a problem of WG prediction. Logistic regression, MLPs and C4.5 classification trees and CART algorithms are tested in a problem of WG prediction at Kumeu, New Zealand.
In \cite{shanmuganathan2014data} a similar problem was tackled, also in New Zealand. In this case, the study evaluates the performance of classification trees, MLPs and Self-Organizing Maps (SOM). In-situ measurements and data acquired between 2008 and 2012 at Kumeu site, have been used for this study.
In \cite{lagerquist2017machine} a problem of extreme wind prediction at the surroundings of storm cells in the USA is carried out. The problem consists in calculating the probability of extreme winds over 50kt (25.7 m/s) in zones close to storm cells. The problem is formulated as a binary classification problem. The predictive variables considered in this case are based on radar measurements, storm motion and shape, and atmospheric soundings at the near-storm environment. Several ML models have been tested, including, logistic regression, RF, MLPs and Gradient boosting trees ensembles.
In \cite{wang2020probabilistic} an ensemble model for WG prediction is presented. The proposed ensemble includes RF, a long-short term memory (LSTM) algorithm and Gaussian processes for regression. A comparison against each model on their own, the persistence and a gradient boosted decision tree showed the good performance of the ensemble method.
Also dealing with ensemble models, in \cite{schulz2021machine} a comprehensive review and comparison of eight ensemble methods based on ML for WG forecasting is carried out. The proposed algorithm are tested in six years of data from a high-resolution ensemble prediction system of the German weather service.
In \cite{spassiani2021application} a SOM is proposed to analyze the meteorological origin of WG in Australia. The SOM is used to establish the origin of Application of Self-organizing Maps to classify the meteorological origin of WG into convective (from thunderstorms) and non-convective origin (synoptic), with different subclasses in each case.

Finally, in \cite{arul2022machine} a RF approach is applied to the identification of extreme wind field characteristics and associated wind-induced load effects on structures, via detection of thunderstorms. The idea is to use large databases containing high-frequency sampled continuous wind speed data, and use the shapelet transform to identify individual attributes distinctive of extreme wind events. Experiments using real data from 14 Mediterranean ports, including sites in Italy, Spain and France are carried out.

\subsubsection{Analysis}
The large majority of EEs related to severe weather are meteorological events, in which thermodynamics processes of the atmosphere plays a central role. Depending on the EEs considered as severe weather, the period of return of the EEs is extremely high, such as damaging hailstorms, though other EEs classified as severe weather are much more frequent. Techniques to take into account the physics of these EEs in the ML are based on NWM (the ML algorithms are applied to the output of NWM) such as In \cite{gagne2015day}, as the most effective method to consider the thermodynamic processes that characterize these EE, together with in-situ measurements, such as radar reflectivity or convective indices \cite{gagne2017storm,czernecki2019application}. However this, note that we have classified as severe weather different meteorological events, with specific peculiarities. For example, convective systems and hail storms are related events, quite local, in which thermodynamics and atmospheric state play an important role, very difficult to include as predictive variables in ML approaches. In extreme winds and gust, however, the dynamics of the atmosphere may have significant importance to describe the phenomenon, and thus the synoptic situation provides information which may be exploited by ML algorithms \cite{spassiani2021application}, in addition to other local atmospheric variables describing convective systems.

\subsection{Fog and extreme low-visibility}

Low-visibility EE, usually associated with fog formation \cite{Gultepe07} or turbidity in the atmosphere due to pollution, deeply affect transportation facilities such as airports~\cite{Cornejo20,guerreiro2020analysis} and roads \cite{peng2018analysis,wu2018crash}. ML algorithms have been successfully applied in the last years to many fog and low-visibility prediction problems. 

In \cite{Fabbian07} MLPs were tested in a problem of fog events prediction at Canberra International Airport (Australia), from meteorological observations. Data from the Australian Bureau of Meteorology were used to train and test the neural networks, obtaining promising results.
In \cite{Miao12} a fog prediction system formed by fuzzy logic-based predictors was proposed and analyzed at Perth airport (Australia). The fuzzy logic predictor worked on the outputs of a meso-scale numerical model (LAPS125) outputs, with the objective of refining the predictions obtained by the numerical model. This fog prediction model was operational at the airport and its outcomes averaged with the outcomes of two other fog forecasting methods by means of a majority voting approach.
In \cite{Colabone15} the performance of MLPs with back-propagation training procedure in a fog event prediction problem at Academia da For\c{c}a A\'erea (Brasil) is analyzed.
In \cite{Boneh15} a Bayesian network is applied to a fog prediction problem at Melbourne airport. In this case the problem is tackled as a prediction time horizon of 8-hours, and 34 years of data have been used to train the network. This fog prediction system has obtained better results than previous systems, becoming operational for fog prediction at Melbourne airport.
In \cite{bartokova2015fog} a decision tree for short-time fog prediction in Dubai is presented. The decision tree is able to improve the results of meso-scale models such as WRF in short-term prediction time horizons up to 6 hours.
In \cite{Cornejo17} different ML regression techniques have been tested over a fog prediction problem at Valladolid airport, Spain. In this case, radiation-type fog events are the most common the zone, so the prediction problem is restricted to winter months. The authors reported successful  results in for event prediction by using Support Vector Regression algorithms and Extreme Learning Machines approaches.
In \cite{zhu2017application} a deep neural network has been applied to a problem of low-visibility prediction at Urumqi airport, China. Meteorological variables measured at the airport between 2007 and 2016 are used to feed the deep neural network.
In \cite{Duran18} evolutionary neural networks are considered for a problem of fog events classification from meteorological input variables. Several types of evolutionary neural networks are considered, by selecting different basic neurons types (sigmoidal, product and radial). A multi-objective training procedure is considered, obtaining good results in the fog event classification problem considered.
In \cite{Guijo18} a problem of low-visibility events due to fog is tackled by applying ordinal classification methods. Three classes were considered (fog, mist and no-fog), and different ordinal classifiers successfully tested in this problem of fog event prediction.
In \cite{dietz2019forecasting} decision trees models and tree-based ensemble with boosting are applied to a problem of very short-term prediction of low-visibility procedures states at Vienna airport, Austria. The work shows that for prediction time-horizons under 1 hour, the current low-visibility state (persistence), cloud ceiling, and horizontal visibility are the most important variables to take into account. For longer prediction time-horizons visibility information at the airport's surroundings and meteorological variables become relevant.
In \cite{miao2020application} a Long-Short Term Memory (LSTM) neural network has been applied to a problem of fog forecasting in the Anhui province, China. A comparison with K-Nearest Neighbor, AdaBoost and CNN algorithms have shown that the LSTM network is able to obtain better results.
In \cite{bari2020machine} different ML algorithms (tree-based ensembles, feed-forward neural networks and generalized linear methods) have been applied to the output of a NWM (meso-scale model, WRF), for a problem of low-visibility  prediction in Northern Morocco.
In \cite{li2020diagnostic} a decision tree algorithm (C4.5 approach) has been applied to a problem of low-visibility prediction at Nanjing city. The work has shown that in this case the variables related to humidity and particle concentrations (relative humidity, PM10 and PM2.5) are the most important factors to obtain accurate predictions of visibility at Nanjing. 
Finally, in \cite{yu2021application} a hybrid approach mixing Extreme Gradient boosted and NWM have been applied to a problem of visibility prediction at Shanghai, China. A large number of predictive variables are considered such as air pollutants concentration, meteorological observations, aerosol optical depth data and satellite images. The proposed hybrid approach provides a more accurate visibility forecast for prediction time horizons of  24 and 48h than LGBM algorithms and NWM on its own.

In \cite{cornejo2020persistence} the persistence and ML prediction of low-visibility events are studied in Valladolid airport, Spain. The performance of binary classifiers is evaluated in a problem of radiation for prediction in winters. In \cite{cornejo2021statistical} a problem of low-visibility events prediction due to orographic forcing is analyzed with ML regressors at Lugo, Nothwestern Spain. The work includes the statistical analysis of the low-visibility events in this zone.
In~\cite{CASTILLOBOTON2022106157} a thorough comparison of several ML algorithms in fog prediction problems is carried out. Both classification and regression techniques are analyzed, including balancing techniques and augmented data methods to improve the performance of ML in fog events prediction.
In close connection with low-visibility events, in this case due to storms, in \cite{ebrahimi2021predicting} the number of dusty days is predicted with ML techniques in Northern Iran. SVR, RF and Stochastic Gradient boosting are the ML algorithms successfully applied in this problem. In \cite{ding2022forecast} the prediction of hourly low-visibility events is tackled in 47 Chinese airports, by means of different ML approaches such as MLP, RF, regression trees (CART) and KNN approaches, among others. The results obtained show important differences in performance from different airports, and also at different seasons (better performance in the cold season than in the warm season).

\subsubsection{Analysis}
ML analysis of fog events has been intense in the last years. Fog formation may follow different physical mechanisms \citep{Gultepe07}. For example, radiation fog, a typical fog of inland areas, usually occurs in winter under anticyclonic conditions, when clear skies and stability of the atmosphere allows the nocturnal radiative cooling required to saturate the air \citep{Roman-Gascon12}. On the other hand, advection fog occurs when moist, warm air passes over a colder surface and is cooled from below, producing a immediate condensation of water. This kind of fog is very common at sea, when moist and unstable warm air moves over cooler waters. If the moist warm air moves up to a hill or slope, the air undergoes an adiabatic expansion which, in turn, cools down the air as it rises, allowing the moisture in it condensing and this way producing fog, usually called orographic or hill fog. Note that the dissipation mechanisms and persistence of these fog events are also different depending on the formation process \cite{Cornejo20,salcedo2021long}. Inclusion of physics in ML approaches should take into account these formation and dissipation mechanisms, depending on the type of fog event considered. The best way of taking into account this is to consider as inputs meteorological variables related to fog formation or dissipation, as in the majority of cases has been done. Also, there have been some works which have used NWM as a previous step before the application of ML algorithms, as in \cite{bari2020machine,yu2021application}.

\subsection{Final discussion}

As reviewed in previous sections, a large amount of ML algorithms have been applied to a wide class of problems in EEs detection, prediction and attribution. EEs problems in different spatio-temporal scales have been tackled with ML algorithms. In some cases, long-term physical processes related to the atmospheric dynamics seem to be predominant (heatwaves, extreme temperatures, droughts and floods in some cases), while in other cases, local short-term processes associated with thermodynamics is the predominant factor of the problem (convective systems, flash floods or extreme fog events). 

We have broadly detected three types of approaches using ML in the literature reviewed, in all EEs problems considered in this work. First, there are articles in which ML algorithms have been applied {\em raw}, i.e. without any reference to the physics related to the problem. Usually these works proposed approaches based on time series of measured values, or involved some signal processing techniques, such as series decomposition, wavelets, etc. In general these approaches have been exclusively compared against other alternative methods fully based on ML or autoregressive approaches such as ARIMA methods, and a poor discussion on the physical reasons of the good or badly performance of the algorithms is carried out. A second type of approaches described in the literature reviewed are those works which try to take into account the physics of the problem through the input variables considered in the ML methods. Depending on the problem considered, certain input variables may consider physical aspects of the problems, such as atmospheric dynamics (synoptic situations, Rossby waves, climate indices and other variables related to atmospheric dynamics) or thermodynamics processes (convective or stability indices, and other variables related to thermodynamics process, usually from reanalysis data, satellites or direct measurements). Finally, the third type of works revised in this section are those ML approaches which present hybridization with physical or numerical models considering the physics of the problem, or those which present a coupling with physical models in order to improve their outputs. Different versions including hybridization/coupling with numerical models such as WRF, Analogue-based algorithms, and other NWM have been revised in this section. In general, these latter hybrid approaches were successfully compared with physical models and also with other ML approaches. In some cases, future projections based on CMIP6 models have been carried out from ML approaches, in attribution-related problems.

It is also remarkable the fact that different problem encodings and frameworks have been used in the EEs problems revised. Classification and regression frameworks have been used, depending on the specific EE, at very different spatio-temporal scales, from local to synoptic and global scales, at short-term and long-term temporal scales. The number of input variables in ML algorithms is an important issue in many of the approaches revised. In many cases FS mechanisms are needed in order to improve the results of ML algorithms. In general, the articles reviewed reported successful ML applications to EE, but the comparison with alternative approaches can be biased. For example, those approaches in which physics process are not taken into account in the ML, are not usually compared to alternative approaches including physical models, but only with other ML methods. In those works in which ML methods have been hybridized with NWM to include the physics of the EE, an improvement over the NWM has been reported. In many cases, this ML hybridization with NWM is focused on downscaling processes, in order to improve the spatial resolution of NWM, by using ML algorithms.


\section{Case study: summer temperature prediction with ML approaches. Results and open problems} \label{sec:SumTemPred}

ML approaches devoted to characterize and predict heatwaves and extreme temperature have been previously discussed in this paper (Section \ref{sec:heatwaves}). In this case study, different problem formulations are shown and discussed, also some results and issues related to summer temperatures prediction, where heatwaves signals can be detected, based on reanalysis data for France. A final subsection shows an outlook, findings summary and open problems from this case study.

\subsection{August mean temperature prediction in France based on ML approaches and synoptic predictive variables from reanalysis}

\begin{figure}[!ht]
    \centering
    \includegraphics[draft=false, angle=0,width=12cm]{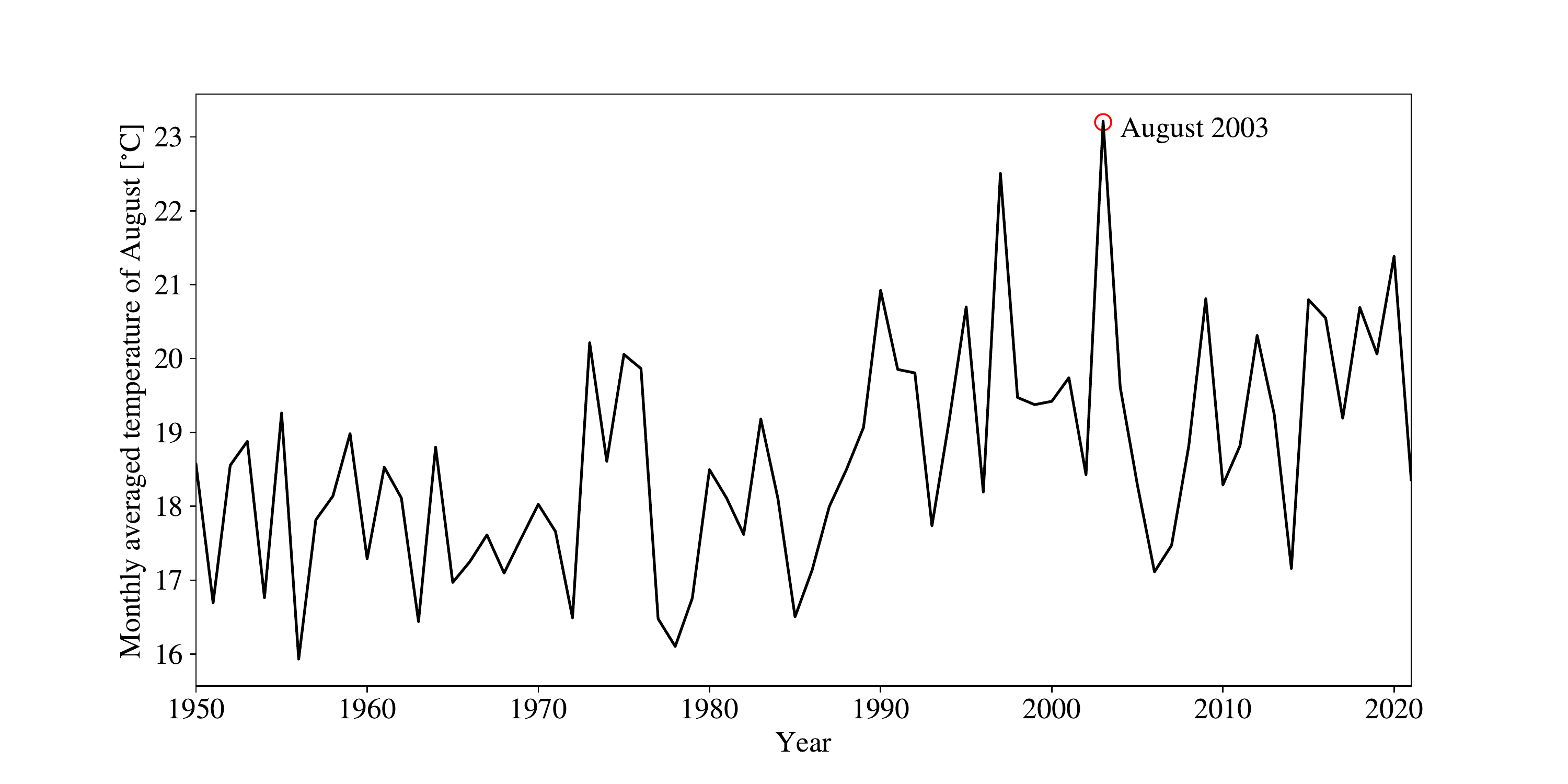}
    \caption{\label{fig:Paris_1950to2021} Monthly averaged temperature in Paris, between 1950 to 2021. Mega-heatwave of 2003 is highlighted in the time series.}
\end{figure}

In this first problem definition, the prediction of August mean temperature by using ML approaches is approached. In order to give a first definition of the problem, a specific case of August mean temperature prediction in central France is considered, where there have been extremely hard summer heatwaves in the last 20 years \cite{garcia2010review,ouzeau2016heat,barriopedro2011hot}. Let $T(t)$ an objective time series of air temperature (2~m temperature, for instance, or any other similar air temperature variable), obtained at a given point or averaged over a set of known points. In our case, $T(t)$ stands for the mean temperature of a summer month (August) in the location of interest. Air temperature from ERA5 reanalysis data \cite{hersbach2020era5} has been considered in this case. Figure \ref{fig:Paris_1950to2021} shows the objective August mean temperature (2m temperature) in Paris area (France) from 1950 to 2021. Note that in some cases it is possible to spot heatwaves signal in $T(t)$, such as the mega-heatwave of August 2003 in Europe.

\begin{figure}[!ht]
    \centering
    \includegraphics[draft=false, angle=0,width=14cm]{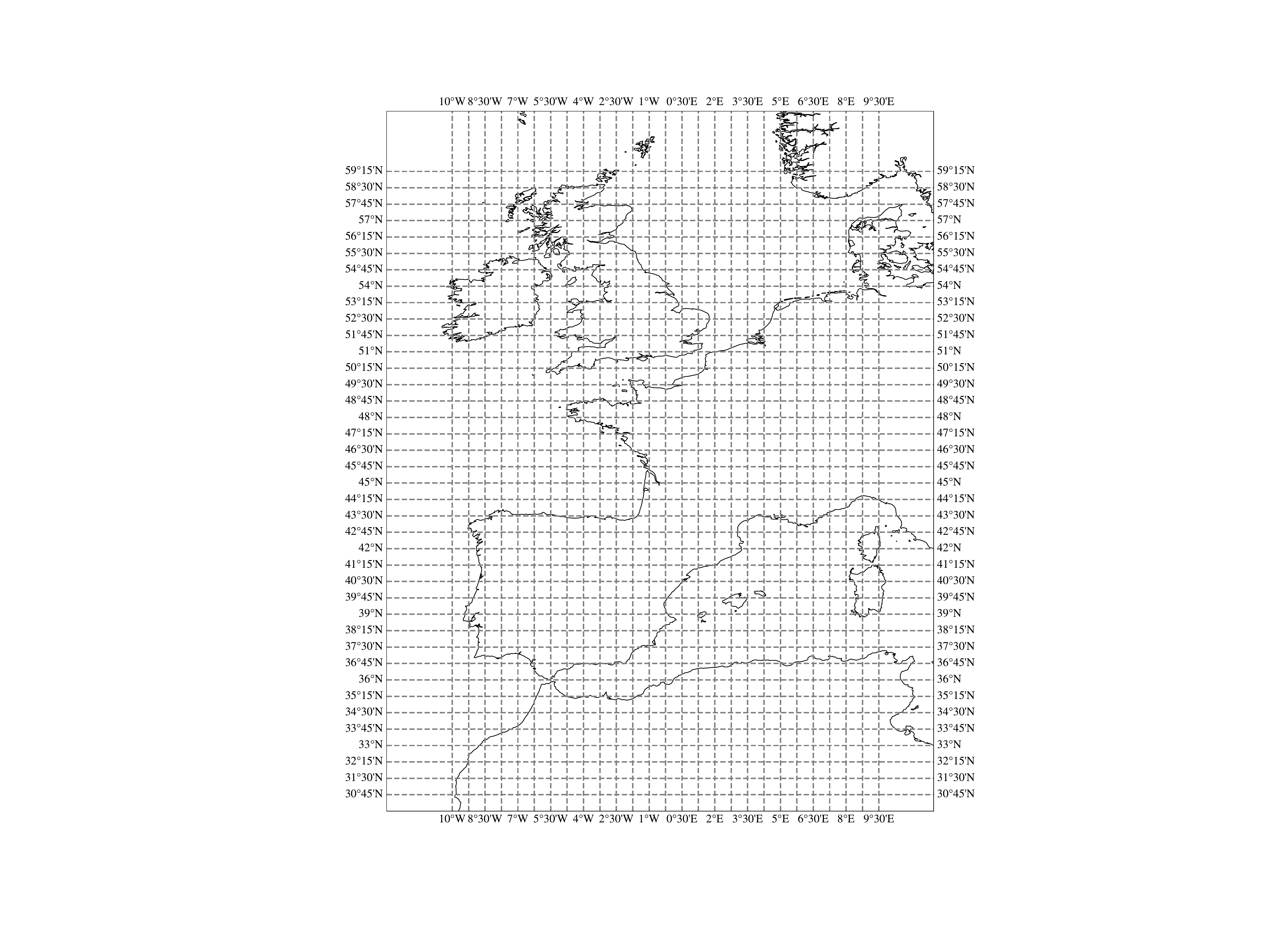}
    \caption{\label{fig:grid} Synoptic regular grid similar to the one considered in this case study.}
\end{figure}

Let $V(t',{\bf x})$ be the set of predictive variables, usually defined in a spatial regular grid $\bf x$, over time. Note that we have notated $t'$ since it may not match with time $t$ in $T(t)$. In this first problem, we consider a synoptic regular grid (Figure \ref{fig:grid}), covering France, where we define a number of predictive variables to estimate $T(t)$, also obtained from ERA5 reanalysis \cite{hersbach2020era5}. Table \ref{tab:variables_description} shows the predictive and target variables considered in this work. 

\begin{figure}[!ht]
    \centering
    \includegraphics[draft=false,trim={0 5cm 0 5cm}, angle=0,width=10cm]{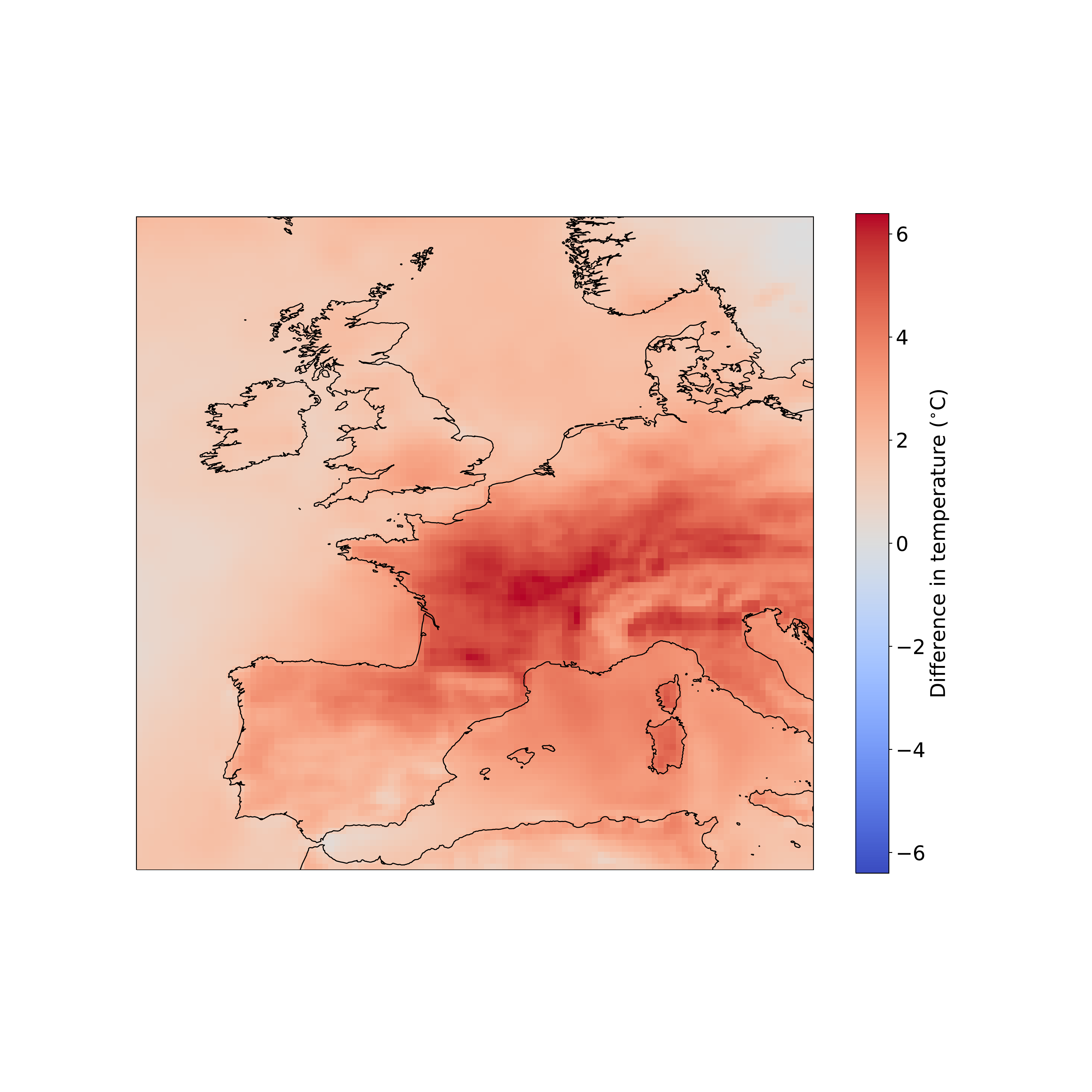}
    \caption{\label{fig:difference_temperature} Anomaly of monthly averaged temperature in August between 2003 and the averaged temperature from 1950 to 2002.}
\end{figure}

\begin{table}[htpb]
\centering
\caption{ Predictive and target variables considered in this case study, as obtained by \cite{ecmwf}.}
\begin{tabular}{cl}
\hline\hline
 Short name in ERA5 & Description \\ \hline\hline
 u10m       & Eastward component of wind at a height of 10m \\ 
 v10m       & Northward component of wind at a height of 10m \\
 msl & Mean sea level pressure\\ 
 swvl1  & Volumetric soil water at layer 1 (0-7~cm)           \\
 t2m & Air temperature at 2~m (target)\\
 \hline\hline
\end{tabular}
\label{tab:variables_description}
\end{table}

We consider the problem of predicting the mean temperature of August $T(t)$ (regression problem), by using the value of the predictive variables in the previous months ($t'$ stands for months of July/June, same year) in $V(t',x)$. This approach is similar to that in \cite{Oettli2022}, but focused on the summer temperature. Different ML and DL techniques among those described in Section \ref{sec:methods} are considered to tackle this problem. Specifically,  RF, DT, MLP, SVR and LSTM networks. We have also included a Linear Regression approach for comparison purposes.

Several research questions arise here: for instance, we want to assess whether there is enough information from variables in $V(t',x)$ to obtain a good quality prediction of $T(t)$ from ML approaches. Regarding extreme values, we want to know if we are able to obtain a ML prediction mechanism which shows a good quality prediction of extreme temperature values with a prediction time-horizon of one month in advance. Also, the problem of obtaining the best set of features for the ML algorithms arises here. In order to solve these research questions, we will show different results and we will discuss different open problems found when dealing with this case study. 

\subsection{Experimental results and research issues}

We have structured the results obtained in several subsections, where we discuss results considering input variables from one single reanalysis node (local approach), results from several reanalysis nodes (synoptic approach), issues regarding the prediction problems, mainly the number of training samples available, and how to solve them by including new training samples with oversampling approaches.

\subsubsection{Input variables from one reanalysis node}

\begin{figure}[!ht]
    \centering
    \includegraphics[draft=false, angle=0,width=12cm]{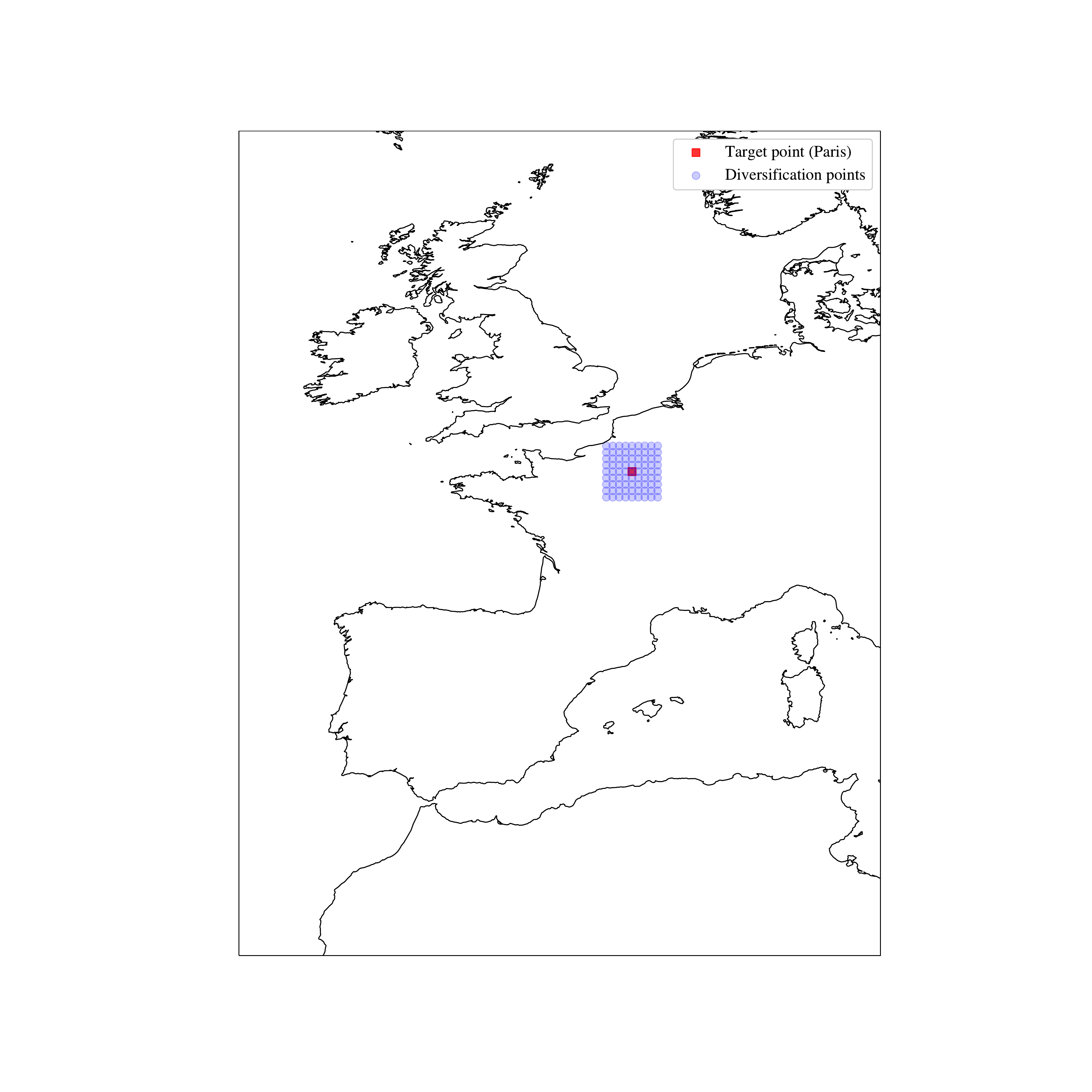}
    \caption{\label{fig:regions} Mean August temperature prediction problem considered. Reanalysis nodes for predictive (blue) and objective variables (red).}
\end{figure}

\begin{figure}[!ht]
    \centering
    \includegraphics[draft=false, angle=0,width=12cm]{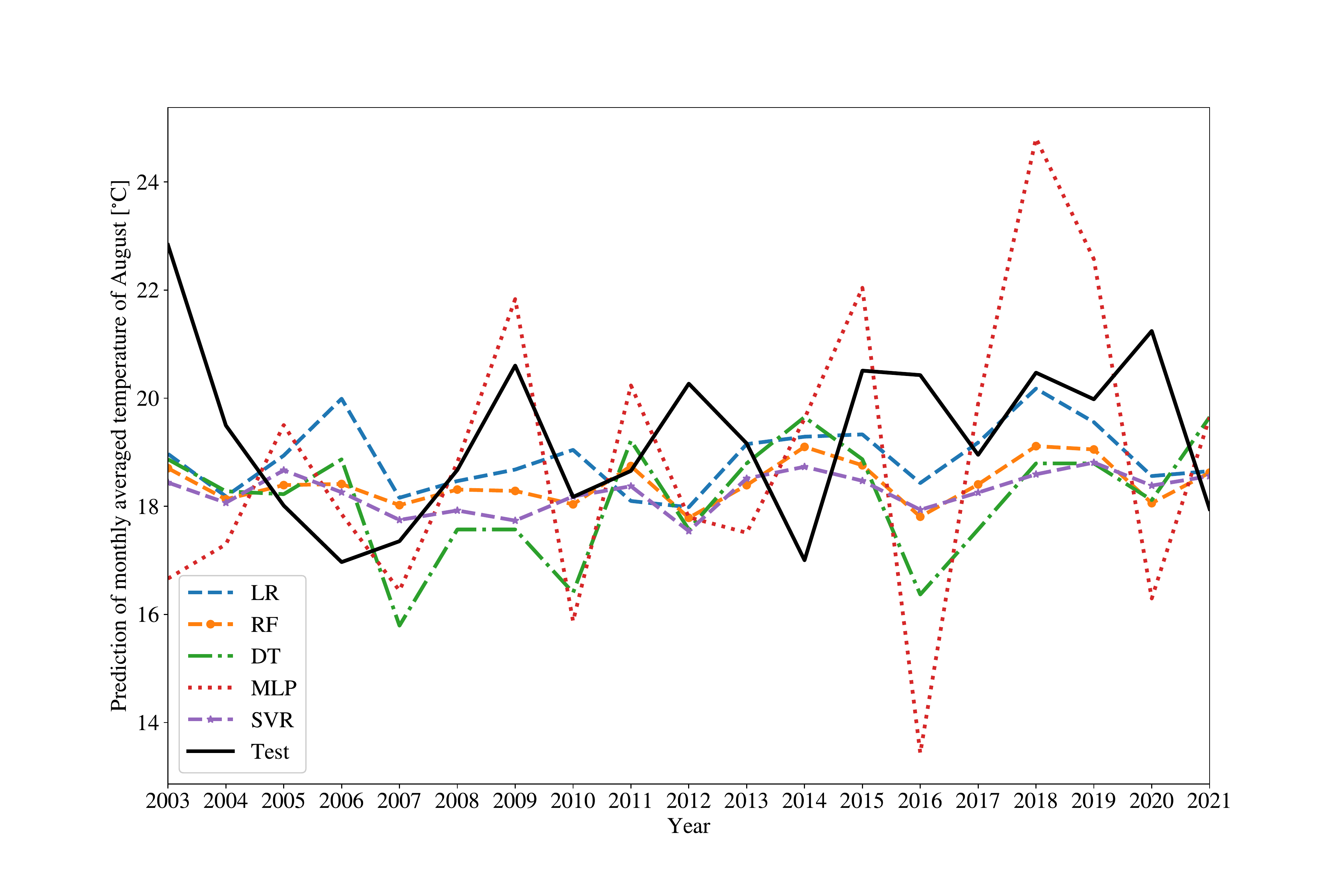}
    \caption{\label{fig:local_single_point} Temperature prediction considering variables from a single reanalysis node.}
\end{figure}

We start with a simple regression problem in one single node of the reanalysis field, to evaluate a first approach to the problem. Let us consider then one single node in France, where we collect the predictive variables, and try to predict the target (August temperature), from them. Figure \ref{fig:regions} shows the considered node, in red. In order to tackle the problem, we first consider a training and test partition of the data. We have available data from 1950 up to 2021, and we consider the period 1950-2002 for training, and the period 2003-2021 as the test set to evaluate the results. Note that, since we consider annual data (August temperature and predictive variables in July), we only have 53 samples for training the algorithms, and we consider 19 test samples where we evaluate the performance of the ML approaches. Table \ref{tab:local_pred_metrics} (first column) shows the MAE obtained by the different ML algorithms for this simple first case, and Figure \ref{fig:local_single_point} details the predictions obtained by each ML algorithm. As can be seen, the prediction obtained by the ML algorithms is in general not fully accurate. Note, however, that some of the ML approaches are able to estimate $T$ with more accuracy than others. MLP seems to be the worst approach with a MAE of 2.45, DT obtains the second worst result, with a MAE of 1.88, whereas LR, RF and SVR seem to work better, with MAE values of 1.34, 1.43 and 1.52, respectively. If we examine the regression problem tackled, we can see that we count on 53 samples, with 5 predictive variables each to train the ML algorithms. The results evidence that the amount of information from the predictive variables with only 53 samples is clearly not enough to get a extremely good prediction result. 

\begin{table}[!ht]
	\centering
	\caption{MAE and MSE of ML regressors considered in the heatwave prediction problem with by reanalysis spatial diversity.}
	\begin{tabular}{c|cc}
		\hline\hline
		& \multicolumn{2}{c}{Single Point} \\ \hline\hline
		Method.                 &  MAE     & MSE    \\ \hline
    	LR       &  1.34      & 2.94     \\ 
		RF  &  1.43       & 3.25          \\ 
		DT                &  1.88                & 4.71  \\ 
		MLP &  2.45      & 9.30        \\ 
		SVR                  &  1.52     & 3.55         \\ \hline\hline
	\end{tabular}
	\label{tab:local_pred_metrics}
\end{table}

\subsubsection{Exploiting spatial diversity of reanalysis data to improve ML accuracy}

The question that arises at this point is: \emph{can we obtain or generate somehow more training samples in order to improve the prediction of heatwaves with ML algorithms?} In this section we show a possible strategy to generate additional training samples (oversampling procedure), by exploiting spatial diversity of reanalysis data.

Let us get back to the problem tackled above, with a single reanalysis node considered, and 5 predictive variables. If we consider a local approach, note that there are a large number of reanalysis nodes in the neighborhood of the selected one. In Figure \ref{fig:regions} we have set a number of neighbor reanalysis nodes in blue (81 nodes), around the red point. It is important to note that we have all the predictive (input) and objective ($T$) variables in all the points considered. Since we are in a local approach, we can assume a similar behavior of the variables in the selected grid, in such a way that we can use all the variables in the grid as training samples. Note that this way we are introducing an oversampling approach \cite{torgo2015resampling}, by exploiting the diversity of reanalysis in a local approach. In this particular case, we finally obtain 4293 training samples ($81 \times 53$) instead of the initial 53 samples.

\begin{table}[htpb]
	\centering
	\caption{MAE and MSE of ML regressors considered in the mean August temperature prediction problem in France, with and without oversampling by reanalysis spatial diversity.}
	\begin{tabular}{c|cc}
		\hline\hline
		                      &  \multicolumn{2}{c}{Oversampling} \\ \hline\hline
		Method.                 &  MAE        & MSE \\ \hline
    	LR       &  1.35       & 2.93\\ 
		RF       &  1.50           & 3.26\\ 
		DT       &  1.02         & 1.46\\ 
		MLP      &    1.54        & 3.25\\ 
		SVR      &  1.36      & 3.28\\ \hline\hline
	\end{tabular}
	\label{tab:local_pred_metrics2}
\end{table}

\begin{figure}[!ht]
    \centering
    \includegraphics[draft=false, angle=0,width=12cm]{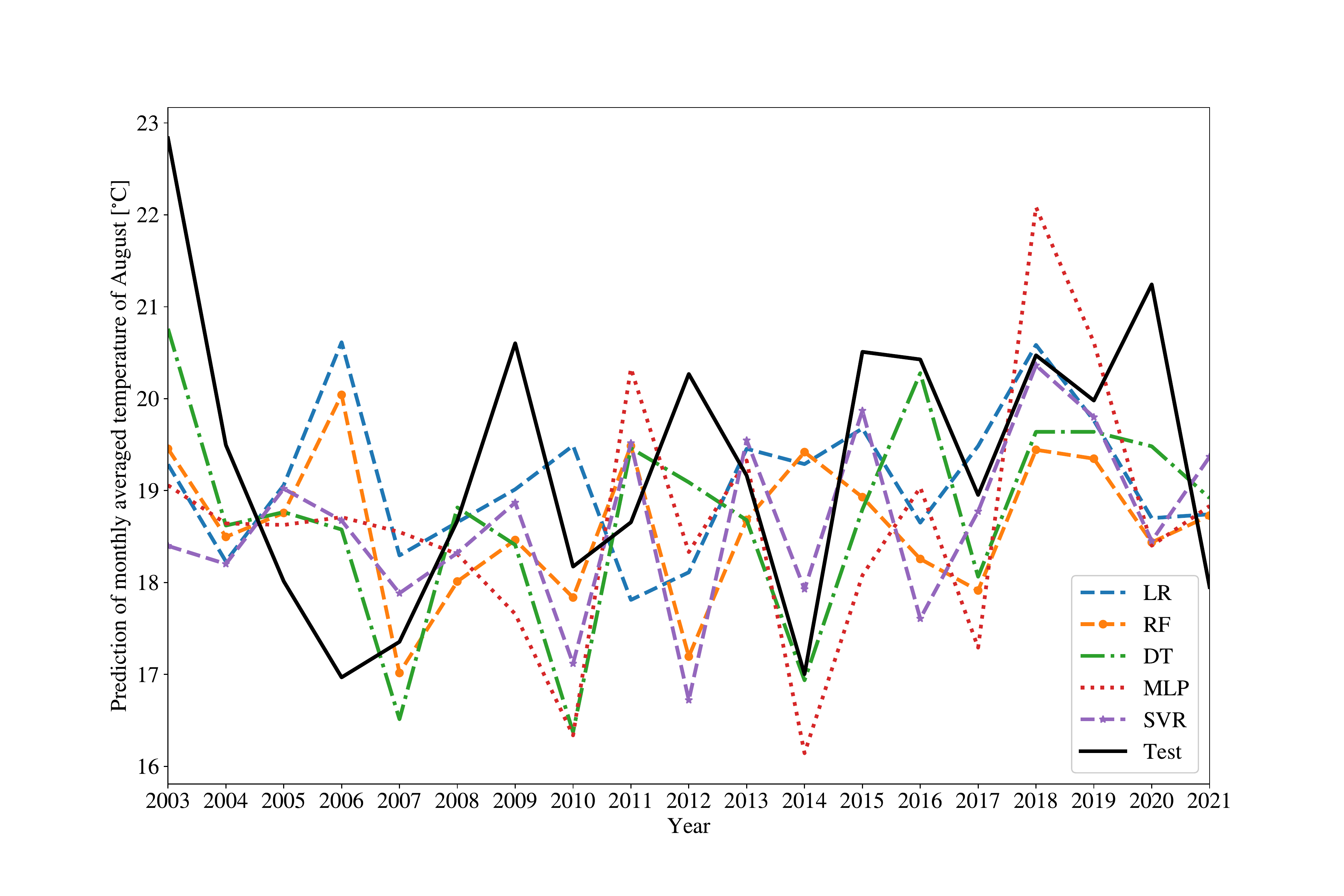}
    \caption{\label{fig:local_multiple_points} Temperature prediction considering oversampling by reanalysis spatial diversity (one initial reanalysis node).}
\end{figure}

Table \ref{tab:local_pred_metrics2} and Figure \ref{fig:local_multiple_points} show the new results when reanalysis spatial diversity is included to generate oversampling. As can be seen, we obtain much better prediction results in the majority of cases by considering a larger training set. The best improvement is for DT (MAE 1.88 $\rightarrow$ MAE 1.02) an excellent result with accurate prediction. The MLP is also much improved with the new oversampling by reanalysis spatial diversity (MAE 2.45 $\rightarrow$ MAE 1.54), and the SVR is also improved in this case (MAE 1.52 $\rightarrow$ MAE 1.36). The LR and RF do not improve their result when oversampling by reanalysis diversity is considered, but their performance deterioration is not very accused.

This way, we have shown how oversampling by considering reanalysis spatial diversity is able to improve the results obtained by the ML regressors in the temperature prediction problem considered.

\subsubsection{Extension to several input reanalysis nodes}

\begin{figure}[!ht]
    \centering
    \includegraphics[draft=false, angle=0,width=12cm]{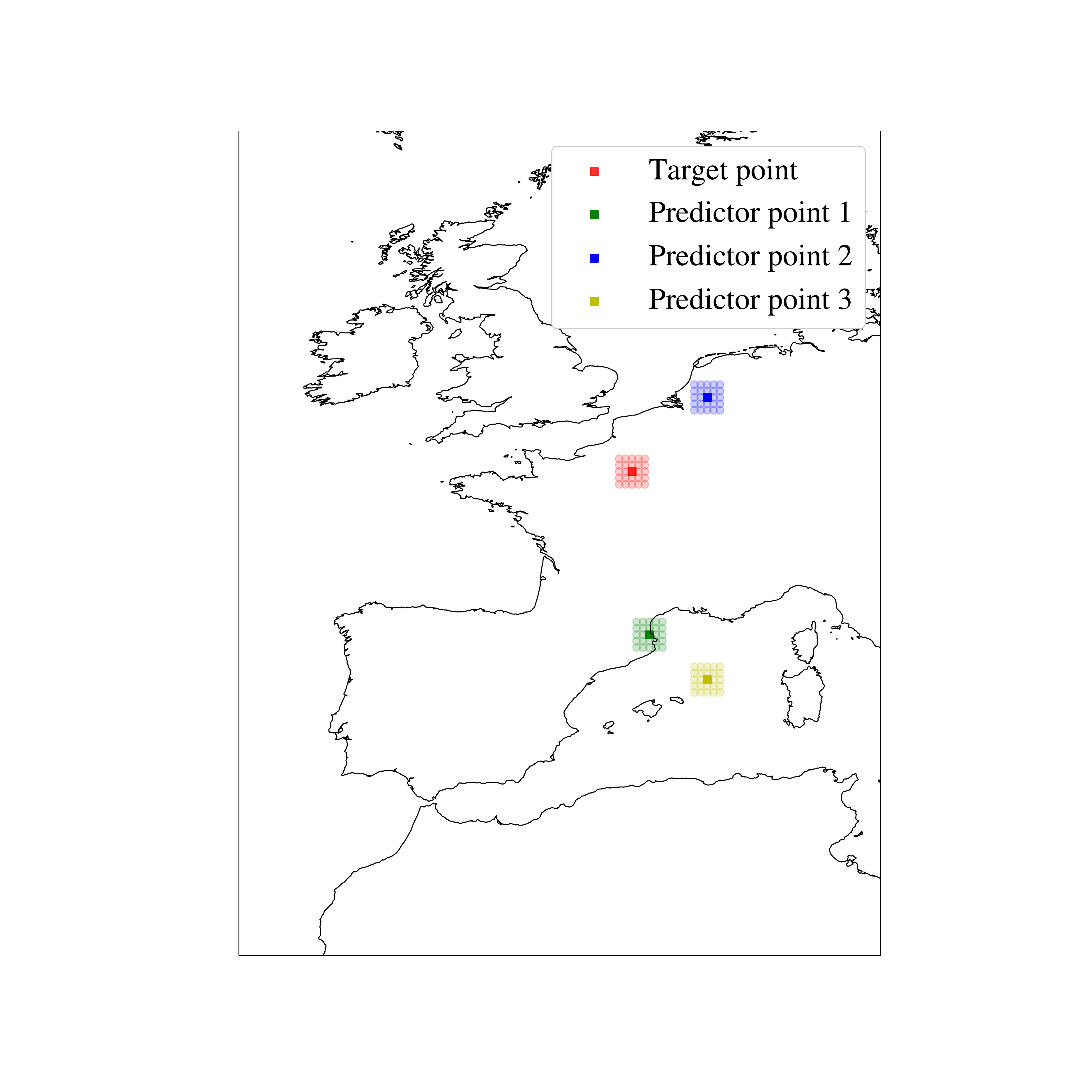}
    \caption{\label{fig:general_div} Temperature prediction problem with input variables from 4 reanalysis nodes.}
\end{figure}

Let us consider a second problem, with several reanalysis nodes to carry out the prediction of the heatwaves. We show a case with four reanalysis nodes in Figure \ref{fig:general_div}.

Note that, in this case, we consider 5 variables per node of reanalysis, so adding a node implies adding 5 more predictive variables to the prediction. Thus, a total of 20 predictive (input) variables are now considered in the problem, with 53 training samples in this case. It is expected that increasing the number of input variables with just 53 training samples  do not lead to better results, due to the increasing in the number of predictive (input) variables (20 in this case). Table \ref{tab:local_pred_metrics3} shows the results obtained with all the ML regressors considered. As can be seen, the prediction of $T(t)$ is not better now than that considering a single point of reanalysis, or the improvement is very little, as in the case of MLP and SVR algorithms.

\begin{table}[htpb]
	\centering
	\caption{MAE and MSE from ML regressors considered in the heatwave prediction problem with four input reanalysis nodes.}
	\begin{tabular}{c|cc}
		\hline\hline
		                      &  \multicolumn{2}{c}{4 Reanalysis nodes} \\ \hline\hline
		Method.  &  MAE        & MSE \\ \hline
    	LR       &  2.27      & 6.60\\ 
		RF       &  1.44      & 3.04\\ 
		DT       &  1.73      & 3.90\\ 
		MLP      &  2.07      & 7.45\\ 
		SVR      &  1.49      & 3.30\\ \hline\hline
	\end{tabular}
	\label{tab:local_pred_metrics3}
\end{table}

\begin{figure}[!ht]
    \centering
    \includegraphics[draft=false, angle=0,width=12cm]{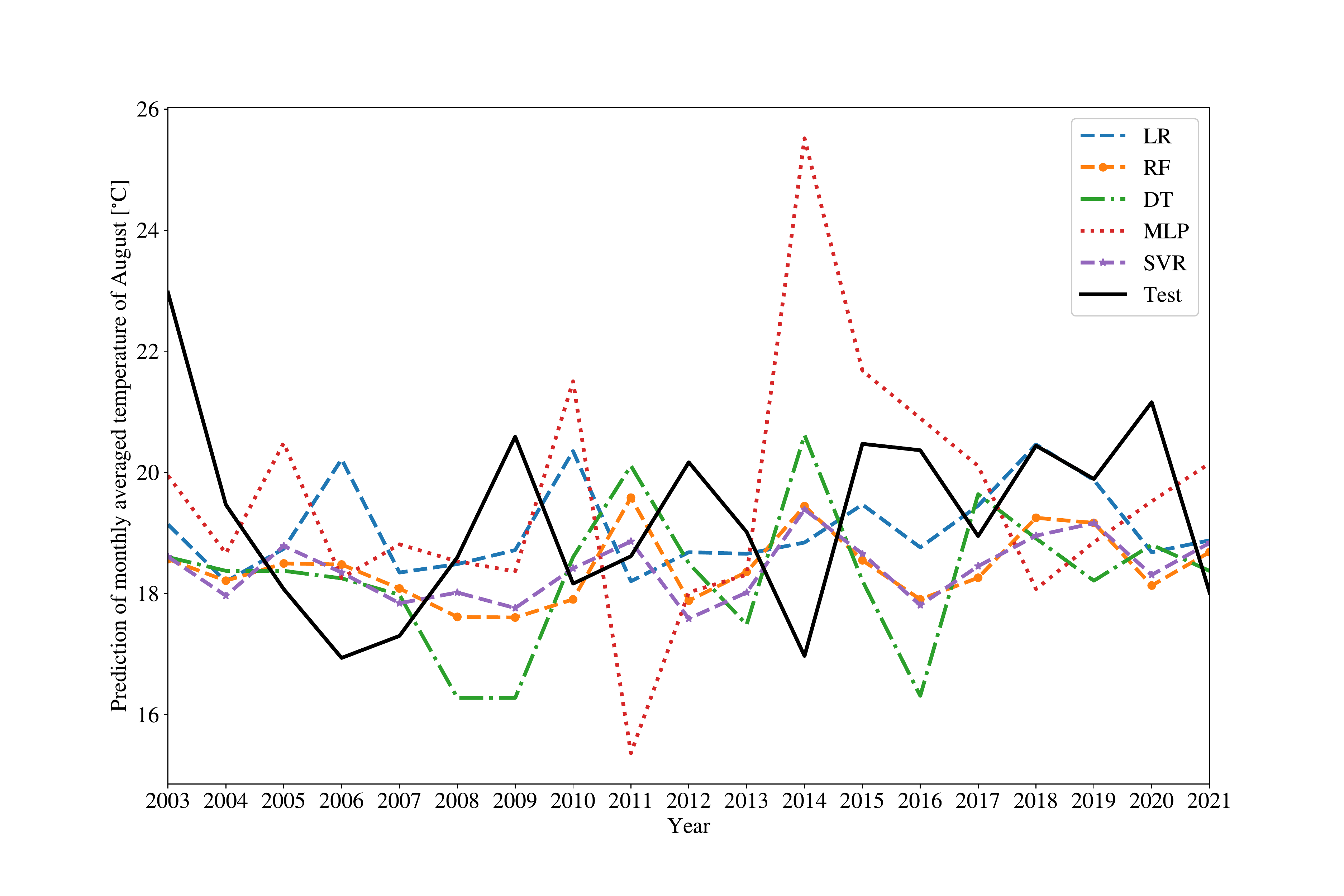}
    \caption{\label{fig:general_NPoints} General case with diversification.}
\end{figure}

We can also introduce oversampling by exploiting ERA5 spatial diversity data when we consider several reanalysis nodes to predict $T(t)$. Figure \ref{fig:general_div} shows the diversity generation in this case. Note that for each reanalysis node, we can generate diversity by randomly selecting a neighbor node in each one. This way we can exploit the fact that the neighbor reanalysis nodes provide similar predictive variables or target values, and we can generate a large number of new training samples. Table \ref{tab:local_pred_metrics4} shows the results obtained by including oversampling by reanalysis spatial diversification. As can be seen, the LR improves a lot its result, and the rest of ML algorithms seem to be slightly affected by diversification in this case, obtaining slightly worse results in general. Figure \ref{fig:general_NPoints} shows the results obtained in the test set, which are, as can be seen, worse than those obtained by considering a single reanalysis node with oversampling.

\begin{table}[htpb]
	\centering
	\caption{MAE and MSE of ML regressors considered in the mean August temperature prediction problem in France, with and without oversampling by reanalysis spatial diversity.}
	\begin{tabular}{c|cc}
		\hline\hline
		                      &  \multicolumn{2}{c}{Oversampling(4 nodes)} \\ \hline\hline
		Method.  &  MAE        & MSE \\ \hline
    	LR       &  1.30      & 2.82\\ 
		RF       &  1.56      & 3.61\\ 
		DT       &  1.90      & 5.27\\ 
		MLP      &  2.07      & 7.45\\ 
		SVR      &  1.53      & 3.56\\ \hline\hline
	\end{tabular}
	\label{tab:local_pred_metrics4}
\end{table}



\subsubsection{ML-based oversampling and undersampling approaches}

In ML, an {\em oversampling procedure} consists of increasing the number of observations by generating new data samples, in order to improve the performance of the training algorithms. In a classification problem, it is common the use of oversampling techniques in unbalance problems, so the oversampling procedure increases the number of observations of all the under-represented classes. There are different oversampling techniques, depending on the task considered (classification or regression). In classification, the most commonly used algorithm is the SMOTE algorithm~\cite{chawla2002smote}, which creates new samples taking into account the statistical of existing ones, diminishing the risk of creating samples in ``wrong'' areas. In the case of regression there are also oversampling techniques that have been recently introduced in the literature such as the SMOGN algorithm \cite{branco2017smogn}.

On the contrary, undersampling methods decrease the number of observations by removing samples. It is also used in unbalanced classification problems, by decreasing the majority classes. As it happens with oversampling, undersampling has to reduce the number of samples while also maintaining the statistical properties of the classes, so it should remove, when possible, redundant information.

\begin{figure}[!ht]
    \centering
    \includegraphics[draft=false, angle=0,width=12cm]{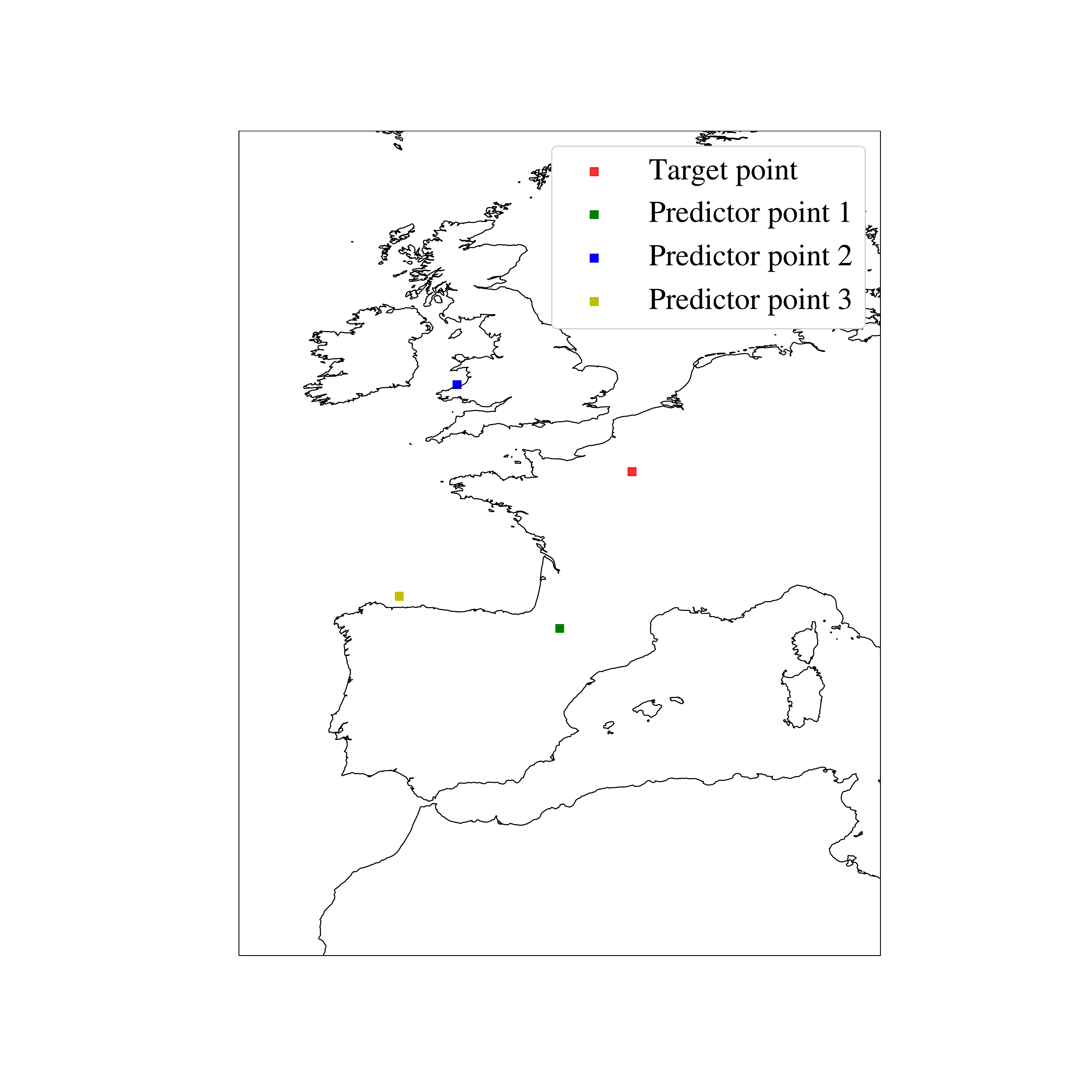}
    \caption{\label{fig:general_div_smogn} Mean August temperature prediction problem in France, with input variables from 4 reanalysis nodes in the case in which SMOGN method is considered.}
\end{figure}

\begin{figure}[!ht]
    \centering
    \includegraphics[draft=false, angle=0,width=12cm]{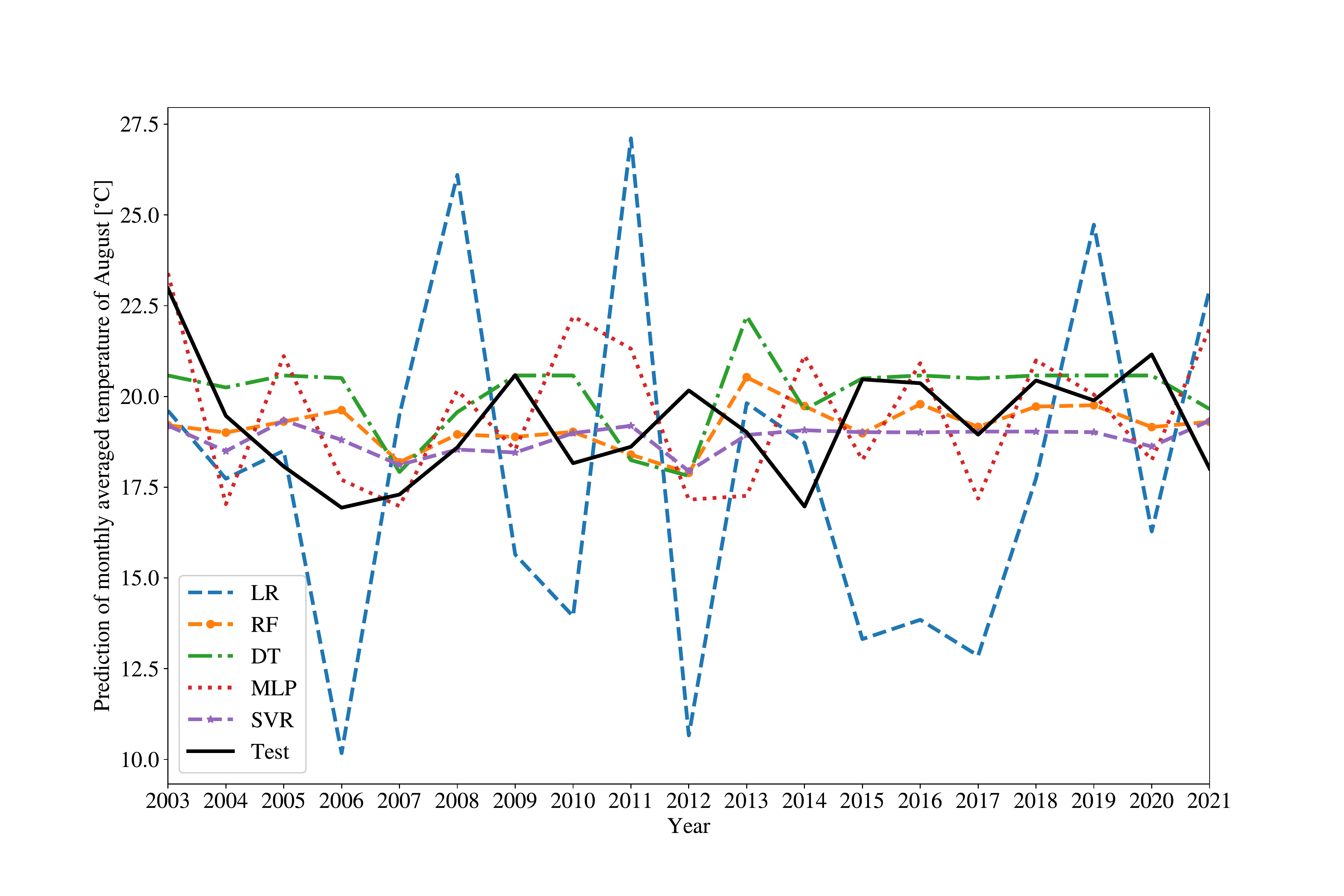}
    \caption{\label{fig:general_div_with_smogn} Temperature prediction by applying SMOGN oversampling method. }
\end{figure}

In order to test ML-based oversampling approaches, we consider another case with four reanalysis nodes (Figure \ref{fig:general_div_smogn}), and the SMOGN algorithm to generate ML-based oversampling in the problem. Tables \ref{tab:local_pred_metrics_NOSMOGN2} and \ref{tab:local_pred_metricsSMOGN2} show the results obtained without considering oversampling by SMONGN and considering it. As can be seen, the performance of all tested regressors is improved by considering oversampling, but the LM, which considerably worsens its results without oversampling. Figure \ref{fig:general_div_with_smogn}.

\begin{table}[htpb]
	\centering
	\caption{MAE and MSE of ML regressors considered in the temperature prediction problem without and with SMOGN technique applied on the database  by reanalysis spatial diversity.}
	\begin{tabular}{c|cc}
		\hline\hline
		 &  \multicolumn{2}{c}{4 Reanalysis nodes} \\ \hline\hline
		Method. &  MAE    & MSE   \\ \hline
    	LR      &  2.13     & 6.95\\ 
		RF      &  1.38     & 2.86\\ 
		DT      &  1.77     & 4.24\\ 
		MLP     &  2.01     & 5.67\\ 
		SVR     &  1.51     & 3.30\\ \hline\hline
	\end{tabular}
	\label{tab:local_pred_metrics_NOSMOGN2}
\end{table}

\begin{table}[htpb]
	\centering
	\caption{MAE and MSE of ML regressors considered in the mean August temperature prediction problem in France, without and with SMOGN technique applied on the database by reanalysis spatial diversity.}
	\begin{tabular}{c|cc}
		\hline\hline
		 &  \multicolumn{2}{c}{SMOGN applied} \\ \hline\hline
		Method. &  MAE    & MSE  \\ \hline
    	LR      &  4.67       & 28.40\\ 
		RF      &  1.32       & 2.71\\ 
		DT      &  1.40       & 3.23\\ 
		MLP     &  2.01       & 5.67\\ 
		SVR     &  1.35       & 2.67\\ \hline\hline
	\end{tabular}
	\label{tab:local_pred_metricsSMOGN2}
\end{table}


\subsection{Case study outlook, findings and open problems}

In this case study we have discussed the application of ML algorithms to a problem of mean temperature prediction in August from reanalysis data in France. We have defined the problem in this way in order to extend it to the prediction of heatwaves in France, when smaller spatio-temporal scales are considered. In fact, even at monthly scale, a heatwave signal can be detected in August mean temperature in some cases of meta-heatwaves, as that of 2003 in France \cite{garcia2010review}. In addition, we have observed the following issues from the application of the ML algorithms:

\begin{itemize}
    \item Since the problem definition involves annual samples (temperature in August), the training set has very small number of samples. This point, combined with the fact that we have a large grid with a large number of predictive variables on it, makes the training of the ML an extremely hard task.
    
    \item The results obtained in a first problem considering a single reanalysis node, are far from accurate, due to the scarce number of training samples.
    
    \item In order to improve the performance of the ML approaches, we propose to exploit the spatial diversity of the reanalysis data considered. First, we consider a fully local approach, including in the training set a number of neighbor reanalysis nodes to the objective node to generate new training samples. This oversampling approach generates new training samples, which allows a better training of the ML algorithms, improving the results obtained in the prediction of August mean temperature.
    
    \item  In a second attempt, we consider several reanalysis nodes to make the prediction, and oversampling by exploiting local diversity in each node. The prediction obtained in these two cases by the ML algorithms are poorer than in the previous cases, since the number of predictive variables is increased, and much more training samples would be necessary to improve the results. 
    
    \item We have also shown the performance of ML algorithms in this problem, by considering ML-based oversampling by applying the SMOGN algorithm. SMOGN is specially suited for regression problems. In this problem of August mean temperature prediction in France the SMOGN works fine, producing oversampling which improves the performance of all ML algorithms versus the case without oversampling.
    
    \item Thus, we have shown that considering oversampling to expand the training set is a good option in this prediction problem with scarce number of data. We have proposed a novel oversampling approach by exploiting the spatial diversity of Reanalisys data, and we have also shown that ML-based oversampling also works in the problem. 
    
\end{itemize}

There are several open problems in the prediction of annual mean air temperature from reanalysis data. We summarize them in the following points:

\begin{itemize}
    \item We have tackled local and synoptic versions of the problem from reanalysis data, with predictive variables back to just one month before. However, it is known that heatwaves detection (signal in mean monthly temperature) may have different drivers, some of them related with climate indices, which points out to a global definition of variables, with time-horizon for these predictive variables back to several months before. In this global definition of the problem, the management of the huge number of features involved will be extremely important to obtain significant results. Also, the generation of enough training samples for the ML algorithms is again a challenging aspect of the global version of the problem.

    \item Note that there are different possible definitions of this prediction problem, depending on the data considered. We have shown an example with monthly temperature data, but quarterly, weekly or even daily time precision can be chosen and will also contain heatwave signal. It is also possible to directly use heatwaves indices \cite{awasthi2022retrospection,nairn2015excess} to define the problem, which have been proposed in the past, including other variables in addition to temperature.
    
    \item In close relationship with the latter point, note that the problem can be tackled as a regression or classification problem. We have shown here a regression version, where the direct prediction of $T(t)$ is tackled. In a classification problem, $T(t) \rightarrow s[n]$, where $s[n] \in \{0,1\}$ if we consider a binary classification problem (heatwave signal detected/no heatwave signal). This problem can be extended to a larger number of classes.
    
    \item We have shown how the problem cannot be successfully tackled without considering the physics of the phenomenon. In other words, the ML approaches must be coupled with the physics of extreme temperatures, which act at different levels and considering different physical aspects of the problems, such as atmospheric dynamics and thermodynamics processes in order to improve the quality of the prediction.
    
    \item There are also open problems related to the ML algorithms. As previously mentioned, it is clear that feature selection (see Section \ref{FS_methods}) is key for ML approaches to obtain significant results in the different versions of the problem. Due to the huge number of features involved in the problem, it is probable that wrapper methods on their own do not lead to good results, and a first feature discarding process based on filter approaches is needed. Other possible solutions such as using clustering approaches to reduce the number of features in some specific zones can also provide good results when the number of features is huge, such in the global approach to the problem. 
    
    \item Deep Learning (DL) approaches could be used to tackle the problem without taken special care about the huge number of features involved. In this approach, DL algorithms could be useful to exploit global information and obtain an accurate prediction of heatwaves. Issues related to DL training, such as the number of training samples, and significance of the results obtained are the counterpart of this possible approach with DL algorithms.
    
\end{itemize}

\section{Conclusions and perspectives for future research}  \label{sec:Conclusions}

In this paper we have carried out a review of ML methods in the analysis, characterization, prediction and attribution
of extreme atmospheric events (EEs). It is currently a hot topic, since EEs are increasing in the current situation of climate change, causing important damages to human and ecosystems. After a brief review of the main ML approaches which have been previously applied to EE-related problems, we have carried out a comprehensive and critical analysis of this topic in the literature, including the main EE, such as extreme rainfall and floods, heatwaves and extreme temperatures, droughts, fog and low-visibility events, and different topics related to severe weather (convective systems, tropical cyclones, hailstorms and extreme winds). 

We have shown the application of several ML methods to a case study related to mean summer temperatures prediction in France, from reanalysis (ERA5) data. We have shown the main issues related to this problem using ML, including the scarce number of samples to train the ML approaches, the huge number of input variables and the different possible problem's definitions (regression or classification tasks, prediction time-horizon considered, etc.). We have also shown that the inclusion of the physics is a key point in order to obtain good results for this problem, so it is necessary to couple the ML algorithms with some physical information for the problem in order to improve the results obtained.

Note that these issues associated with the case study considered in this paper can be extrapolated to other similar problems in extreme atmospheric events, which share similar data structure and scarce of events and data. We have also given some solutions to these issues for the case study considered, such as including oversampling techniques from reanalysis diversity, or even using different reanalysis data or global climate models to generate new training samples for the ML algorithms. These proposed solutions can also be applied to other problems related to extreme atmospheric events. 

\subsection{Perspectives}  \label{sec:Perspectives}
We also discuss here some final lessons learned, open problems and research possibilities and direction which are currently an option for dealing with EEs using ML algorithms, such as the use of XAI techniques, improving the attribution of EEs with ML techniques, and improving the study of concurrent and compound events, where the lack of data to train the ML algorithms is even more pressing.

\begin{itemize}
\item One of the main issue when dealing with ML approaches to EEs prediction problems are the databases. Given the rarity of EEs, there are very few long-enough databases which provide reliable data for EE-Related studies. Even reanalysis data, with more than 70 years of data world-wide with high spatial accuracy may be not enough for some problems (the case study presented before is a good example of this). In theses cases, oversampling data may be of great help to improve the performance of ML algorithms. Note that only by considering two different reanalysis data (ERA5 and ERA20C, for example \cite{salcedo2020machine}), we can duplicate the number of samples in the training set, by considering the output of each reanalysis in the same nodes. This opens the possibility to use climate models (with different parameterizations) to multiply the number of training samples available. Another interesting possibility is the application of different oversampling techniques to increase the number of training samples in a given database. In the case of reanalysis-type data, or data defined in a regular grid, oversampling can be carried out in a natural way by considering neighbor nodes, or with tailor-made techniques depending on the specific problem considered. Yet, the use of model-based data (either reanalysis or climate models' simulations) could potentially limit the ability of ML methods of learning relationships outside the ones already implemented in the model. Moreover, training a ML algorithm on model-based data could overestimate the performance when tested against observational data as model-based data do not perfectly reproduce the real climatic conditions due to modeling errors and assumptions \cite{Hoffmann2020drought,matsuoka2022obs}. We therefore advocate making the most of observational data as they represent a richer ground truth, although sometime characterized by low data quality and missing values. Here, however, ML can also contribute with advanced methodologies to reconstruct missing climate information \cite{kadow2020artificial}.

\item Another niche of opportunity in the characterization of EEs is the use of explainable AI techniques to gain an informed understanding of the correlations modeled by ML models \cite{arrieta2020explainable}. Indeed, a large fraction of the ML models used nowadays in this area rely on complex structures and processing units (e.g. deep neural networks) that achieve unrivaled levels of performance at the cost of an opaque training and inference processes. This clashes with other models which, by virtue of their transparent internal structure or the way they are trained, elicit interpretable information about what features are relevant for the target at hand (e.g. tree-based bagging ensembles or linear regression). When this interpretability is not provided off-the-shelf, explanations can be generated ad-hoc for already trained models producing, as a result, visualizations, quantitative scores of predictive relevance or alternative what-if hypothesis for the model's input, to mention a few \cite{montavon2018methods}. This growing concern with explanatory techniques for ML models has spawned a whole area of research coined as eXplainable Artificial Intelligence (XAI), becoming a topic of central importance in applied machine learning in almost any discipline.


Very recently, such techniques have started to be explored for extreme events prediction, as early as 2022. This is the case of \cite{van2022using}, where XAI was used to verify that a ML model learned to predict high summer temperatures from multiple predictors at different time scales agrees with the theoretical understanding of the underlying physical processes. However, there still prevail several challenges that, in our vision, should congregate the efforts of the community in years to come. Among them, we highlight two differential research directions: 

\begin{enumerate}
\item The need for stepping beyond correlation-based ML towards data-based causality inference \cite{peters2017elements}. Since the goal of decision making is to avoid -- or at least, minimize the consequences of -- extreme events, data-based models should guarantee the actionability of the model's input to steer the predicted output in one direction or another. Such interventional tools are being actively investigated nowadays in the context of ML, with models ensuring input-output causality still far from their maturity (see \cite{runge2019inferring} and references therein) because they often require the introduction of several assumptions (e.g. Gaussian distributions) that might be violated by the processes associated with EEs. At the same time, less assumptions are required for dentifying the absence of a causal link \cite{runge2018causal}, making the findings of noncausality already quite robust in determining when it is unlikely that a cause-effect physical mechanism exists. We expect the use of data-based causality inference to become more and more attractive for supporting the trustability of black-box ML models \cite{reichstein2019deep}.

\item The inherent uncertainty of the physical world and the atmosphere propagates to the output of the models devised to characterize extreme phenomena occurring therein. Thereby, a remarkable corpus of literature has striven towards quantifying the confidence of the model in its output considering the modeling (epistemic) uncertainty and the irreducible (aleatoric) uncertainty. While confidence analysis is a well-established area in ML research, the combination of confidence and explainability in a single framework is still to be seen. Indeed, explanations of uncertain models makes no practical sense, nor do models that are certain about their predictions but do not explain what they model in the data at hand. The variability and incompleteness of atmospheric data, and the large epistemic uncertainty of Deep Learning models can, with no doubts, leverage advances such as evidential DL, variational neural networks or model-agnostic techniques such as conformal prediction. Confidence estimations provided by these techniques should be considered when furnishing explanations.
\end{enumerate}

On a summarizing note, we advocate for a focus of the research community steered towards modeling aspects that complement the derivation of more models and performance comparison studies. In other words, we advocate for ML approaches at the end of a pipeline driven by physics, as in this review we have shown very different examples which show that the application of ML techniques without including the physical basis of the problem does not lead to relevant results in the majority of cases.   

\item Improving attribution of EEs using ML. There are not many works dealing with attribution of EEs using ML techniques. In this work we have discussed some works dealing with attribution of EEs using ML techniques for specific events of heatwaves \cite{pasini2017attribution} and droughts \cite{richman20182015}. There are some recent works dealing with ML in general climate attribution problems \cite{mamalakis2021neural}, and also on specific attribution of forced climate change signals over atmospheric fields such as global temperature or precipitation \cite{barnes2019viewing,barnes2020indicator,hartigan2020attribution}. In \cite{callaghan2021machine} a large study on attribution of climate impacts with ML methods has been recently carried out. However, it is necessary to extend these works to better cope with attribution of EEs by using ML approaches. The application of novel ML approaches specific for attribution problems is another line to follow in the years to come. The study of causal inference with ML \cite{scholkopf2019causality} is also a topic fully related to attribution, in which there are some recent works focused on extreme atmospheric events \cite{nethery2020integrated,liu2021method}.

\item ML for concurrent and compound EEs. The concept of {\em concurrent event} refers to (atmospheric) EEs of different types occurring within a specific temporal lag, either in different locations or at the same one. This concept can also be used for extremes of the same type occurring in two different locations within a specific time period \cite{toreti2019concurrent}. On the other hand, {\em compound events} refer to concomitant (within a given temporal lag) occurrence of events (extremes or not) with serious and harmful consequences of socio-economic relevance. It is possible to see that concurrent events are a subset of compound events. In spite of the work on concurrent and compound events has been intense in the last years \cite{bresch2018future,zscheischler2020multivariate,white2021atmospheric,markonis2021rise}, the application of ML techniques to prediction or attribution of concurrent or compound events has been minor. There is a very recent work discussing ML techniques applicable to compound events together with statistical and numerical techniques \cite{zhang2021compound}, and some white papers and technical reports on the topic \cite{feng2021characterization}, but in general the application of ML to this topic is an open problem. The most important issue with ML approaches in concurrent and compound EEs is related to the lack of available data to study these type of situations. There have been some intents to generate databases for concurrent and compound events \cite{feng2020database}, but in general further efforts are needed to strength this topic, so ML methods can be successfully applied in this area.

\end{itemize}

\section*{Acknowledgement}
This research has been partially supported by the European Union, through H2020 Project ``CLIMATE INTELLIGENCE Extreme events detection, attribution and adaptation design using machine learning (CLINT)'', Ref: 101003876-CLINT. This research has also been partially supported by the project PID2020-115454GB-C21 of the Spanish Ministry of Science and Innovation (MICINN). J. Del Ser also acknowledges support by the Basque Government through EMAITEK and ELKARTEK funds (ref. KK-2020/00049), as well as the consolidated research group MATHMODE (IT1294-19).


 \bibliographystyle{elsarticle-num} 
 \bibliography{biblio}





\end{document}